\documentclass{article}
\usepackage{ijcai22}

\usepackage{times}

\usepackage{soul}
\usepackage{url}
\usepackage[hidelinks]{hyperref}
\usepackage[utf8]{inputenc}
\usepackage[small]{caption}
\usepackage{graphicx}
\usepackage{amsmath}
\usepackage{booktabs}
\usepackage{algorithm}
\usepackage{algorithmic}
\usepackage{amsfonts}       
\usepackage{lipsum}
\usepackage{multirow}
\usepackage{color}
\usepackage{enumitem}
\usepackage{textcomp}
\usepackage{microtype}
\usepackage{subfig}
\usepackage{booktabs} 
\usepackage{chngpage}
\usepackage{wrapfig}
\DeclareMathOperator{\diag}{diag}

\newcommand{\mat}[1]{\boldsymbol{#1}}
\renewcommand{\vec}[1]{\boldsymbol{\mathrm{#1}}}
\newcommand{\vecalt}[1]{\boldsymbol{#1}}

\providecommand{\mA}{\ensuremath{\mat{A}}}

\providecommand{\mR}{\ensuremath{\mat{R}}}

\providecommand{\mT}{\ensuremath{\mat{T}}}
\providecommand{\mU}{\ensuremath{\mat{U}}}

\providecommand{\vo}{\ensuremath{\vec{o}}}

\providecommand{\vx}{\ensuremath{\vec{x}}}

\providecommand{\vmu}{\ensuremath{\vecalt{\mu}}}
\providecommand{\vsigma}{\ensuremath{\vecalt{\sigma}}}

\title{Model-Based Offline Planning with Trajectory Pruning}

\author{
	Xianyuan Zhan$^1$, Xiangyu Zhu$^1$, Haoran Xu$^2$
\affiliations
$^1$Institute for AI Industry Research (AIR), Tsinghua University, Beijing, China\\
$^2$JD iCity \& JD Intelligent Cities Research, JD Technology, Beijing, China\\
\emails
zhanxianyuan@air.tsinghua.edu.cn, \{zackxiangyu, ryanxhr\}@gmail.com
}

\begin{document}

\maketitle

\begin{abstract}
The recent offline reinforcement learning (RL) studies have achieved much progress to make RL usable in real-world systems by learning policies from pre-collected datasets without environment interaction. Unfortunately, existing offline RL methods still face many practical challenges in real-world system control tasks, such as computational restriction during agent training and the requirement of extra control flexibility. 
The model-based planning framework provides an attractive alternative.
However, most model-based planning algorithms are not designed for offline settings. Simply combining the ingredients of offline RL with existing methods either provides over-restrictive planning or leads to inferior performance.
We propose a new light-weighted model-based offline planning framework, namely MOPP, which tackles the dilemma between the restrictions of offline learning and high-performance planning. MOPP encourages more aggressive trajectory rollout guided by the behavior policy learned from data, and prunes out problematic trajectories to avoid potential out-of-distribution samples. Experimental results show that MOPP provides competitive performance compared with existing model-based offline planning and RL approaches.
\end{abstract}

\section{Introduction}

Recent advances in offline reinforcement learning (RL) have taken an important step toward applying RL to real-world tasks. Although online RL algorithms have achieved great success in solving complex tasks such as games \cite{silver2017mastering} and robotic control \cite{levine2016end}, they often require extensive interaction with the environment. 
This becomes a major obstacle for real-world applications,
as collecting data with an unmatured policy via environment interaction can be expensive (e.g., robotics and healthcare) or dangerous (e.g., industrial control, autonomous driving). Fortunately, many real-world systems are designed to log or have sufficient pre-collected historical states and control sequences data. Offline RL tackles this challenge by training the agents offline using the logged dataset without interacting with the environment. The key insight of recent offline RL algorithms \cite{fujimoto2018off,kumar2019stabilizing,wu2019behavior,yu2020mopo} is to restrict policy learning stay ``close'' to the data distribution, which avoids the potential extrapolation error when evaluating on unknown out-of-distribution (OOD) samples. 

However, implementing offline RL algorithms on real-world robotics and industrial control problems still faces some practical challenges. For example, many control agents have limited computational resources for policy learning, which require a light-weighted policy improvement procedure. Moreover, industrial control tasks often require extra control flexibility, such as occasionally changing reward signals due to altering system settings or certain devices, and involvement of state-based constraints due to safety considerations (e.g., restrict policy to avoid some unsafe states). Most existing offline RL algorithms need computationally extensive offline policy learning on a fixed task and do not offer any control flexibility. 

Model-based planning framework provides an attractive solution to address the above challenges. The system dynamics can be learned offline based on the prior knowledge in the offline dataset.
The policy optimization can be realized by leveraging model-predictive control (MPC) combined with a computationally efficient gradient-free trajectory optimizer such as the cross-entropy method (CEM) \cite{botev2013cross} or model-predictive path integral (MPPI) control \cite{williams2017model}. 
The planning process also allows easy integration with the change of reward signals or external state-based constraints during operation, without requiring re-training agents as needed in typical RL algorithms.

Most model-based planning methods are designed for online settings.
Recent studies \cite{wang2019exploring,argenson2020model} have borrowed several ingredients of offline RL by learning a behavior cloning (BC) policy from the data to restrain trajectory rollouts during planning. This relieves OOD error during offline learning but unavoidably leads to over-restrictive planning. 
Limited by insufficient expressive power,
behavior policies learned using BC often fit poorly on datasets generated by relatively random or multiple mixed data generating policies. Moreover, restricting trajectory rollouts by sampling near behavior policies also impacts the performance of trajectory optimizers (e.g., CEM, MPPI require reasonable state-action space coverage or diversity in order to find good actions), and hinders the full utilization of the generalizability of the dynamics model.
Dynamic models may learn and generalize reasonably well in some low-density regions if the data pattern is simple and easy to learn. Strictly avoiding OOD samples may lead to over conservative planning which misses high reward actions.

We propose a new algorithmic framework, called  \underline{M}odel-Based \underline{O}ffline \underline{P}lanning with Trajectory \underline{P}running (MOPP),
which allows sufficient yet safe trajectory rollouts and have superior performance compared with existing approaches.
MOPP uses ensembles of expressive autoregressive dynamics models (ADM) \cite{germain2015made} to learn the behavior and dynamics from data to capture better prior knowledge about the system.
To enforce better planning performance, MOPP encourages stronger exploration by allowing sampling from behavior policy with large deviation, as well as performing the greedy max-Q operation to select potentially high reward actions according to the Q-value function evaluated from the offline dataset. At the same time, to avoid undesirable OOD samples in trajectory rollouts, MOPP prunes out problematic trajectories with unknown state-action pairs detected by evaluating the uncertainty of 
the dynamics model. These strategies jointly result in an light-weighted and flexible algorithm that consistently outperforms the state-of-the-art model-based offline planning algorithm MBOP \cite{argenson2020model}, and also provides competitive performance as well as much better control flexibility compared with existing offline RL approaches.




\section{Related Work}
\subsection{Offline reinforcement learning}
Offline RL focuses on the setting that no interactive data collection is allowed during policy learning. The main difficulty of offline RL is the \textit{distributional shift} 
\cite{kumar2019stabilizing},
which occurs when the distribution induced by the learned policy deviates largely from the data distribution. Policies could make counterfactual queries on unknown OOD actions, causing overestimation of values that leads to non-rectifiable exploitation error during training.

Existing offline RL methods address this issue by following three main directions. Most model-free offline RL algorithms constrain the learned policy to stay close to a behavior policy through deviation clipping \cite{fujimoto2018off} or introducing additional divergence penalties (e.g., KL divergence, MMD or BC regularizer) \cite{wu2019behavior,kumar2019stabilizing,fujimoto2021minimalist,xu2021offline}. Other model-free offline RL algorithms instead learn a conservative, underestimated value function by modifying standard Bellman operator to avoid overly optimistic value estimates on OOD samples \cite{kumar2020conservative,liu2020provably,kostrikov2021offline,buckman2020importance,xu2021constraints}. Model-based offline RL methods~\cite{yu2020mopo,kidambi2020morel,zhan2021deepthermal}, on the other hand, incorporate reward penalty based on the uncertainty of the dynamics model to handle the distributional shift issue. The underlying assumption is that the model will become increasingly inaccurate further from the behavior distribution, thus exhibits larger uncertainty. All these algorithms require a relatively intensive policy learning process as well as re-training for novel tasks, which make them less flexible for real-world control systems.

\subsection{Model-based planning}
The model-based planning framework provides a more flexible alternative for many real-world control scenarios. It does not need to learn an explicit policy, but instead, learns an approximated dynamics model of the environment and use a planning algorithm to find high return trajectories through this model. 
Online planning methods such as PETS \cite{chua2018deep}, POLO \cite{lowrey2018plan}, POPLIN \cite{wang2019exploring}, and PDDM \cite{nagabandi2020deep} have shown good results using full state information in simulation and on real robotic tasks. These algorithms are generally built upon an MPC framework and use sample efficient random shooting algorithms such as CEM or MPPI
for trajectory optimization. 
The recent MBOP \cite{argenson2020model} further extends model-based planning to offline setting. MBOP is an extension of PDDM but learns a behavior policy as a prior for action sampling, and uses a value function to the extend planning horizon. The problem of MBOP is that its performance is strongly dependent on the learned behavior policy, which leads to over-restrictive planning and obstructs the full potential of the trajectory optimizer and the generalizability of the dynamics model. In this work, we propose MOPP to address the limitations of MBOP, which provides superior planning while avoids undesirable OOD samples in trajectory rollouts. 




\vspace{-5pt}
\section{Preliminaries}
We consider the Markov decision process (MDP) represented by a tuple as $(\mathcal{S}, \mathcal{A}, P, r, \gamma)$, where $\mathcal{S}$, $\mathcal{A}$ denote the state and action space, $P(s_{t+1}|s_t, a_t)$ the transition dynamics, $r(s_t, a_t)$ the reward function and $\gamma \in [0,1]$ the discounting factor. A policy $\pi(s)$ is a mapping from states to actions. We represent $R=\sum_{t=1}^{\infty} \gamma^{t} r(s_t, a_t)$ as the cumulative reward over an episode, which can be truncated to a specific horizon $H$ as $R_H$. 
Under offline setting, the algorithm only has access to a static dataset $\mathcal{B}$ generated by arbitrary unknown behavior policies $\pi_b$, and cannot interact further with the environment. 
One can use parameterized function approximators (e.g., neural networks) to learn the approximated environment dynamics $f_{m}(s_t, a_t)$ and behavior policy $f_b(s_t)$ from the data.
Our objective is to find an optimal policy $\pi^*(s_t)=$ $\arg\max_{a\in \mathcal{A}} \sum_{t=1}^{H} \gamma^{t} r(s_t, a_t)$ given only dataset  $\mathcal{B}$ that maximizes the finite-horizon cumulative reward with $\gamma=1$.

\vspace{-5pt}
\section{The MOPP Framework}
MOPP is a model-based offline planning framework that tackles the fundamental dilemma between the restrictions of offline learning and high-performance planning. Planning by sampling strictly from behavior policy avoids potential OOD samples. The learned dynamics model can also be more accurate in high-density regions of the behavioral distribution. However, this also leads to over-restrictive planning, which forbids sufficient exploitation of the generalizability of the model as well as the information in the data.
MOPP provides a novel solution to address this problem. It allows more aggressive sampling from behavior policy $f_b$ with boosted variance, and performs max-Q operation on sampled actions based on a Q-value function $Q_b$ evaluated based on behavioral data. This treatment can lead to potential OOD samples, so we simultaneously evaluate the uncertainty of the dynamics models to prune out problematic trajectory rollouts. To further enhance the performance, MOPP also uses highly expressive autoregressive dynamics model to learn the dynamics model $f_m$ and behavior policy $f_b$,
as well as uses the value function to extend planning horizon and accelerate trajectory optimization.

\vspace{-3pt}
\subsection{Dynamics and Behavior Policy Learning}
We use autoregressive dynamics model (ADM) \cite{germain2015made} to learn the probabilistic dynamics model $(r_t, s_{t+1})=f_m(s_t,a_t)$ and behavior policy $a_t=f_b(s_t)$. 
ADM is shown to have good performance in several offline RL problems 
due to its expressiveness and ability to capture non-unimodal dependencies in data \cite{ghasemipour2020emaq}.

The ADM architecture used in our work is composed of several fully connected layers.
Given the input $\vx$ (e.g., a state for $f_b$ or a state-action pair for $f_m$), an MLP first produces an embedding for the input, 
separate MLPs are then used to predict the mean and standard deviation of every dimension of the output.
Let $o_i$ denote the $i$-th index of the predicted output $\vo$ and $\vo_{[<i]}$ represent a slice first up to and not including the $i$-th index following a given ordering. ADM decomposes the probability distribution of $\vo$ into a product of nested conditionals: $p(\vo)=\prod_{i}^{I}p(o_i|\vx, \vo_{[<i]})$. 
The parameters $\theta$ of the model $p(\vo)$ can be learned by maximizing the log-likelihood on dataset $\mathcal{B}$: $L(\theta|\mathcal{B})=\sum_{\vx\in \mathcal{B}}\Big[ \sum_{i=1}^{|\vo|} \log p(o_i|\vx, \vo_{[<i]})\Big]$.

ADM assumes underlying conditional orderings of the data. Different orderings can potentially lead to different model behaviors. MOPP uses ensembles of $K$ ADMs with randomly permuted orderings for dynamics and behavior policy, which incorporates more diverse behaviors from each model to further enhance expressiveness.

\vspace{-2pt}
\subsection{Value Function Evaluation}
Introducing a value function to extend the planning horizon in model-based planning algorithms have been shown to greatly accelerate and stabilize trajectory optimization in both online \cite{lowrey2018plan} and offline \cite{argenson2020model} settings. We follow this idea by learning a Q-value function $Q_b(s_t, a_t)$ using fitted Q evaluation (FQE) \cite{le2019batch} with respect to actual behavior policy $\pi_b$ and $\gamma'<1$:
\begin{equation} 
\scalebox{0.97}{$
	\label{eq:fqe}
	\begin{split}
	&Q_b^k(s_i,a_i) = \arg\min_{f\in F}\frac{1}{N}\sum_{i=1}^{N}\big[f(s_i,a_i) - y_i\big]^2 \\
	y_i &= r_i + \gamma' Q_b^{k-1}(s_{i+1}, a_{i+1}),\, (s_i, a_i, s_{i+1}, a_{i+1})\sim \mathcal{B}
	\end{split}
	$}
\end{equation}
A corresponding value function is further evaluated as $V_b(s_t)=\mathbb{E}_{a\sim\pi_b}Q(s_t, a)$.
This provides a conservative estimate of values bond to behavioral distribution.
MOPP adds $V_b$ to the cumulative returns of the trajectory rollouts to extend the planning horizon. This helps shorten horizon $H$ needed during planning. Besides, MOPP uses $Q_b$ to perform the max-Q operation and guide trajectory rollouts toward potentially high reward actions.

%
%
%
%


\vspace{-1pt}
\subsection{Offline Planning}
MOPP is built upon the finite-horizon model predictive control (MPC) framework. 
It finds a locally optimal policy and a sequence of actions up to horizon $H$ based on the local knowledge of the dynamics model. At each step, the first action from the optimized sequence is executed. In MOPP, we solve a modified MPC problem which uses value estimate $V_b$ to extend the planning horizon:
\begin{equation} 
\label{eq:mpc}
\pi^*(s_0)=\arg\max_{a_{0:H-1}}\mathbb{E}\Big[\sum_{t=0}^{H-1}\gamma^t r(s_t,a_t) + \gamma^{H} V_b(s_{H})\Big]
\end{equation}
Obtaining the exact solution for the above problem can be rather costly, instead, we introduce a new guided trajectory rollout and pruning scheme, combined with an efficient gradient-free trajectory optimizer based on an extended version of MPPI \cite{williams2017model,nagabandi2020deep}.

\vspace{3pt}
\noindent\textbf{Guided trajectory rollout.}
The key step in MOPP is to generate a set of proper action sequences to roll out trajectories that are used by the trajectory optimizer.
Under offline settings, such trajectory rollouts can only be performed with the learned dynamics model $f_m$. Using randomly generated actions can lead to large exploitation errors during offline learning. 
MBOP uses a learned behavior policy as a prior to sample and roll out trajectories. 
This alleviates the OOD error but has several limitations.
First, 
the learned behavior policy could have insufficient coverage on good actions in low-density regions or outside of the dataset distribution. This is common when the data are generated by low-performance behavior policies. Moreover, the dynamics model may generalize reasonably well in some low-density regions if the dynamics pattern is easy to learn. Strictly sampling from the behavior policy limits sufficient exploitation of the generalizability of the learned dynamics model. Finally, the lack of diversity in trajectories also hurts the performance of the trajectory optimizer.

MOPP also uses the behavior policy to guide trajectory rollouts, but with a higher degree of freedom. Let $\vmu^{a}(s_t)=[\mu_1^a(s_t),\cdots,\mu_{|\mathcal{A}|}^a(s_t)]^T$, $\vsigma^{a}(s_t)=[\sigma_1^a(s_t),\cdots,\sigma_{|\mathcal{A}|}^a(s_t)]^T$ denote the mean and standard deviation (std) of each dimension of the actions produced by the ADM behavior policy $f_b(s_t)$. MOPP samples and selects an action at time step $t$ as:
\begin{equation} 
\scalebox{0.98}{$
	\label{eq:sample}
	\begin{split}
	&a_t^i \sim \mathcal{N}\bigg(\vmu^{a}(s_t),  \diag \Big(\frac{\sigma_M}{\max \vsigma^{a}(s_t)}\cdot \vsigma^{a}(s_t) \Big)^2\bigg)\\
	&  \mA_t=\{a_t^i\}_{i=1}^{m},\; \forall i\in\{1,\dots,m\},\;t\in \{0,\dots,H-1\}\\
	&  \hat{a}_t = \arg\max_{a\in A_t} Q_b(s_t, a),\; \forall t\in \{0,\dots,H-1\}
	\end{split}
	$}
\end{equation}
where $\sigma_M > 0$ is the std scaling parameter. We allow it to take larger values than $\max \vsigma^a$ to enable more aggressive sampling.
In MBOP, the actions are sampled by adding a very small random noise on the outputs of a deterministic behavior policy, which assumes uniform variance across different action dimensions. 
By contrast, MOPP uses the means $\vmu^a$ and std $\vsigma^a$ boosted by $\sigma_M$ to sample actions ($\vmu^a$, $\vsigma^a$ from the ADM behavior policy $f_b$). This allows heterogeneous uncertainty levels across different action dimensions while preserves their relative relationship presented in data.


We further perform the max-Q operation on the sampled actions based on $Q_b$ to encourage potentially high reward actions. Note that $Q_b$ is evaluated entirely offline with respect to the behavior policy, which provides a conservative but relatively reliable long-term prior information. MOPP follows the treatment in PDDM and MBOP that mixes the obtained action $\hat{a}_t$ with the previous trajectory using a mixture coefficient $\beta$ to roll out trajectories with the dynamics model $f_m$. This produces a set of trajectory sequences $\mT=\{T_1, \dots, T_N\}$, with $T_n=\{(a_t^n,s_t^n)\}_{t=0}^{H-1}, n\in \{1,\dots,N\}$.

\vspace{4pt}
\noindent\textbf{Trajectory pruning.}
The previously generated trajectories in $\mT$ may contain undesirable state-action pairs that are out-of-distribution or have large prediction errors using the dynamics model. Such samples need to be removed, but we also want to keep OOD samples at which the dynamics model can generalize well to extend the knowledge beyond the dataset $\mathcal{B}$. The uncertainty quantification method used in MOReL \cite{kidambi2020morel} provides a nice fit for our purpose, which is evaluated as the prediction discrepancy of dynamics models $f_m^l$, $l\in 1,\dots,K$ in the ensemble: $\text{disc}(s,a) = \max_{i,j}\left\| f_m^i(s,a)-f_m^j(s,a) \right\|_2^2$.

Let $\mU$ be the uncertainty matrix that holds the uncertainty measures $U_{n,t}=\text{disc}(s_t^n,a_t^n)$ for each step $t$ of trajectory $n$ in $\mT$.
MOPP filters the set of trajectories using the trajectory pruning procedure $TrajPrune(\mT, \mU)$. Denote $\mT_p:=\{T_n | U_{n,t}<L, \forall t,n\}$, trajectory pruning returns a refined trajectory set for offline trajectory optimization as:
\begin{equation}
\scalebox{0.9}{$
	\label{eq:prune}
	\begin{split}
	&TrajPrune(\mT, \mU):= \\
	&\begin{cases}
	\mT_p, \qquad\qquad\qquad\qquad\qquad\qquad\qquad\,\;\text{  if } |\mT_p| \geq N_{m}\\
	\mT_p \cup \text{sort} (\mT-\mT_p,\mU)[0:N_{m}-|\mT_p|], \text{ if } |\mT_p| < N_{m}
	\end{cases}
	\end{split}
	$}
\end{equation}
where $L$ is the uncertainty threshold, $N_{m}$ is the minimum number of trajectories used to run the trajectory optimizer (set as $N_m = 0.2 \lfloor N \rfloor$ in our implementation). 
The intuition of trajectory pruning is to remove undesirable state-action samples and produce a set of low uncertainty trajectories.
MOPP first constructs a filtered trajectory set $\mT_p$ that only contains trajectories with every state-action pair satisfying the uncertainty threshold. If $\mT_p$ has less than $N_{m}$ trajectories, we sort the remaining trajectories in $\mT-\mT_p$ by the cumulative uncertainty (i.e. $\sum_t U_{n,t}$ with $T_n \in \mT-\mT_p$). The top $N_{m}-|\mT_p|$ trajectories in the sorted set with the lowest overall uncertainty are added into $T_p$ as the final refined trajectory set.


\vspace{4pt}
\noindent\textbf{Trajectory optimization.}
MOPP uses an extended version of the model predictive path integral (MPPI) \cite{williams2017model} trajectory optimizer that is used similarly in PDDM \cite{nagabandi2020deep} and MBOP \cite{argenson2020model}. MOPP shoots out a set of trajectories $\mT_f$ using the previous guided trajectory rollout and pruning procedure. Let $\mR_f=\{R_1,\dots,R_{|\mT_f|}\}$ be the associated cumulative returns for trajectories in $\mT_f$,
the optimized action is obtained by re-weighting the actions of each trajectory according to their exponentiated returns:
\begin{equation} 
\label{eq:mppi}
A^*_t = \frac{\sum_{n=1}^{|\mT_f|}\exp(\kappa R_n)a_t^n}{\sum_{n=1}^{|\mT_f|}\exp(\kappa R_n)},\; \forall t=\{0,\dots,H-1\}
\end{equation}
where $a_t^n$ is the action at step $t$ of trajectory $T_n\in \mT_f$ and $\kappa$ is a re-weighting factor. 
The full algorithm is in Algorithm \ref{alg1}.

\begin{algorithm}
	\footnotesize
	\caption{Complete algorithm of MOPP}
	\label{alg1}
	\begin{algorithmic}[1]
		\REQUIRE 
		Offline dataset $\mathcal{B}$
		\STATE Train $Q_b$, $K_1$ dynamics models $f_m^l$ and $K_2$ behavior policies $f_b^l$ on $\mathcal{B}$. Initialize $A^*_t=0$, $\forall t\in \{0, \dots, H-1\}$.
		\FOR{$\tau=0...\infty$}
		\STATE Observe $s_\tau$, initialize $\mT,\mR=\emptyset$
		\FOR{$n=1,\dots,N$}
		\STATE $s_0=s_\tau$, $R_n=0$, $T_n=\mathbf{null}$
		\FOR{$t=0\dots H-1$}
		\STATE Sample action $\hat{a}_t$ using $f_b^l(s_t)$ ($l$ randomly picked from $1\dots K_2$) according to Eq.(\ref{eq:sample})
		\STATE $\tilde{a}_{t}=(1-\beta)\hat{a}_t+{\beta} A^*_{t+1}$, ($A^*_{H}=A^*_{H-1}$)
		\STATE Append $(s_t, \tilde{a}_{t})$ into trajectory $T_n$
		\STATE $s_{t+1}=f_m^{l'}(s_t,\tilde{a}_t)^s$, $l'$ randomly picked from $1\dots K_1$
		\STATE $R_n\leftarrow R_n+\frac{1}{K_1}\sum_{k=1}^{K_1}f_m^k(s_t,\tilde{a}_t)^r$
		\STATE 
		$U_{n,t}=\max_{i,j}\left\|{f_m^i(s_t,\tilde{a}_t)-f_m^j(s_t,\tilde{a}_t)}\right\|_2^2$
		\ENDFOR
		\STATE Compute $V_b(s_H)=\sum_{i=1}^{K_Q}Q_b(s_H,a_i)/K_Q$, $\{a_i\}_{i=1}^{K_Q}$ are randomly sampled from  $f_b^l(s_H)$
		\STATE $R_n\leftarrow R_n+ V_b(s_H)$, $\mT\leftarrow \mT \cup \{T_n\}$, $\mR\leftarrow \mR \cup\{R_n\}$
		\ENDFOR
		\STATE Compute $\mT_f=TrajPrune(\mT,\mU)$ according to Eq.(\ref{eq:prune})
		\STATE Update $A^*_t, \forall t=\{0,\dots, H-1\}$ using $\mT_f$ and Eq.(\ref{eq:mppi})
		\STATE Return optimized $a_\tau=A^*_0$
		\ENDFOR
	\end{algorithmic}
\end{algorithm}

\section{Experimental Results}
\label{sec:exp}

\begin{table*}[th]
	\centering
	\scriptsize

		\begin{tabular}{l|l|c|c|c|c}	
			\toprule
			\multicolumn{2}{c|}{} & \multicolumn{2}{|c|}{\textbf{Model-based offline planning methods}} & \multicolumn{2}{|c}{\textbf{Model-based offline RL methods}} \\
			\textbf{Dataset type}    & \textbf{Environment}  & \textbf{MBOP (MBOP $f_b'$)} & \textbf{MOPP (ADM $f_b$)} & \quad\quad\textbf{MBPO}\quad\quad & \textbf{MOPO} \\ 
			\midrule
			random & halfcheetah & 6.3$\pm$4.0 (0.0$\pm$0.0)& 9.4$\pm$2.6 (2.2$\pm$2.2)  & 30.7$\pm$3.9 &\textbf{35.4$\pm$2.5}     \\ 
			random & hopper & 10.8$\pm$0.3 (9.0$\pm$0.2) & \textbf{13.7$\pm$2.5} (9.8$\pm$0.7) &4.5$\pm$6.0 &11.7$\pm$0.4   \\ 
			random & walker2d & 8.1$\pm$5.5 (0.1$\pm$0.0) & 6.3$\pm$0.1 (2.6$\pm$0.1) & 8.6$\pm$8.1 &\textbf{13.6$\pm$2.6}     \\ 
			\midrule
			medium & halfcheetah & 44.6$\pm$0.8 (35.0$\pm$2.5) &\textbf{44.7$\pm$2.6} (36.6$\pm$4.7) &28.3$\pm$22.7 &42.3$\pm$1.6 4      \\ 
			medium & hopper &48.8$\pm$26.8 (48.1$\pm$26.2) &31.8$\pm$1.3 (30.0$\pm$0.8) &4.9$\pm$3.3 &28.0$\pm$12.4     \\ 
			medium & walker2d &41.0$\pm$29.4 (15.4$\pm$24.7)  &\textbf{80.7$\pm$1.0} (15.6$\pm$22.5) &12.7$\pm$7.6 &17.8$\pm$19.3     \\
			\midrule
			mixed & halfcheetah &42.3$\pm$0.9 (0.0$\pm$0.0)  &43.1$\pm$4.3 (32.7$\pm$7.7) &47.3$\pm$12.6  &\textbf{53.1$\pm$2.0}    \\ 
			mixed & hopper &12.4$\pm$5.8 (9.5$\pm$6.9) &32.3$\pm$5.9 (28.2$\pm$4.3) &49.8$\pm$30.4 &\textbf{67.5$\pm$24.7}     \\ 
			mixed & walker2d & 9.7$\pm$5.3 (11.5$\pm$7.3) &18.5$\pm$8.4 (12.9$\pm$5.7) &22.2$\pm$12.7 &\textbf{39.0$\pm$9.6}   \\
			\midrule
			med-expert & halfcheetah &105.9$\pm$17.8 (90.8$\pm$26.9) &\textbf{106.2$\pm$5.1} (37.6$\pm$6.5)  &9.7$\pm$9.5 &63.3$\pm$38.0      \\  
			med-expert & hopper &55.1$\pm$44.3 (15$\pm$8.7)  &\textbf{95.4$\pm$28.0} (44.3$\pm$28.4) &56.0$\pm$34.5 &23.7$\pm$6.0      \\ 
			med-expert & walker2d &70.2$\pm$36.2 (65.5$\pm$40.2) &\textbf{92.9$\pm$14.1} (13.5$\pm$24.2) &7.6$\pm$3.7 &44.6$\pm$12.9    \\ 
			\bottomrule
		\end{tabular}
		
	\vspace{-6pt}
	\caption{Results for D4RL MuJoCo tasks. The scores are normalized between 0 to 100 (0 and 100 correspond to a random policy and an expert SAC policy respectively).
		We report the mean scores and standard deviation (term after $\pm$) of each method. For MBOP and MOPP, we present the scores of the used behavior policies (MBOP $f_b'$ and ADM $f_b$) in the parentheses.
	}
	\label{tab:bench}
	\vspace{-12pt}
	
\end{table*}

We evaluate and compare the performance of MOPP with several state-of-the-art (SOTA) baselines on standard offline RL benchmark D4RL \cite{fu2020d4rl}. We conduct experiments on the widely-used MuJoCo tasks
and the more complex Adroit hand manipulation tasks. 
All results are averaged based on 5 random seeds, with 20 episode runs per seed.
In addition to performance comparison, we also conduct a comprehensive ablation study on each component in MOPP and evaluate its adaptability under varying objectives and constraints.
Due to space limit, detailed experimental set-up, additional ablation on ADM behavior policy and dynamics model, as well as computational performance are included in the Appendix.

\subsection{Comparative Evaluations}
\textbf{Performance on MoJoCo tasks. }
We evaluate the performance of MOPP on three tasks (\texttt{halfcheetah}, \texttt{hopper} and \texttt{walker2d}) and four dataset types (\texttt{random}, \texttt{medium}, \texttt{mixed} and \texttt{med-expert}) in the D4RL benchmark.
We compare in Table \ref{tab:bench} the performance of MOPP with several SOTA baselines, including model-based offline RL algorithms MBPO \cite{janner2019trust} and MOPO \cite{yu2020mopo}, as well as the SOTA model-based offline planning algorithm MBOP.

MOPP outperforms MBOP in most tasks, sometimes by a large margin. It is observed that MBOP is more dependent on its special behavior policy $a_t =f_b'(s_t,a_{t-1})$, which include previous step's action as input. This will improve imitation performance 
under datasets generated by one or limited policies, as the next action may be correlated with the previous action, 
but could have negative impact on high-diversity or complex datasets (e.g., \texttt{random} and \texttt{mixed}). On the other hand, MOPP substantially outperforms its ADM behavior policy $f_b$ especially on the \texttt{med-expert} tasks, which shows  great planning improvement upon a learned semi-performance policy. 

Comparing with model-based offline RL methods MBPO and MOPO, we observe that MOPP performs better in \texttt{medium} and \texttt{med-expert} datasets, but less performant on higher-variance datasets such as \texttt{random} and \texttt{mixed}. 
Model-based offline RL methods can benefit from high-diversity datasets, in which they can learn better dynamics models and apply RL to find better policies.
It should also be noted that training RL policies until convergence is costly and not adjustable after deployment.
This will not be an issue for a light-weighted planning method like MOPP, as the planning process is executed in operation and suited well for controllers that require extra control flexibility. 
MOPP performs strongly in the \texttt{med-expert} dataset, which beats all other baselines and achieves close to or even higher scores compared with the expert SAC policy. 
This indicates that MOPP can effectively recover the performant data generating policies in the behavioral data and use planning to further enhance their performance.

\begin{table}[t]
	\centering
	\scriptsize
		\begin{tabular}{l|c|c|c|c|c|c}	
			\toprule
			\textbf{Dataset} & \textbf{BC}  & \textbf{BCQ}  & \textbf{CQL} & \textbf{MOPO} & \textbf{MBOP} & \textbf{MOPP}  \\ 
			\midrule
			pen-human & 34.4 & 68.9 & 37.5 & -0.6 & 53.4 & \textbf{73.5} \\ 
			hammer-human & 1.5 & 0.5 & 4.4 & 0.3 & \textbf{14.8} & 2.8 \\ 
			door-human & 0.5 & 0.0 & 9.9 & -0.1 & 2.7 & \textbf{11.9} \\ 
			relocate-human & 0.0 & -0.1 & 0.2 & -0.1 & 0.1 & \textbf{0.5} \\ 
			\midrule
			pen-cloned & 56.9 & 44.0 & 39.2 & 4.6 & 63.2 & \textbf{73.2} \\ 
			hammer-cloned & 0.8 & 0.4 & 2.1 & 0.4 & 4.2 & \textbf{4.9} \\ 
			door-cloned & -0.1 & 0.0 & 0.4 & 0.0 & 0.0 & \textbf{5.6} \\ 
			relocate-cloned & -0.1 & -0.3 & -0.1 & -0.1 & \textbf{0.1} & -0.1 \\
			\midrule
			pen-expert & 85.1 & 114.9 & 107.0 & 3.7 & 105.5 & \textbf{149.5} \\ 
			hammer-expert & 125.6 & 107.2 & 86.7 & 1.3 & 107.6 & \textbf{128.7} \\ 
			door-expert & 34.9 & 99.0 & 101.5 & 0.0 & 101.2 & \textbf{105.3} \\ 
			relocate-expert & \textbf{101.3} & 41.6 & 95.0 & 0.0 & 41.7 & 98.0 \\ 
			\bottomrule
		\end{tabular}
	\vspace{-6pt}
	\caption{Results for Adroit tasks. The scores are normalized between 0 to 100 (correspond to a random policy and an expert SAC policy).
	}
	\label{tab:adroit}
	\vspace{-12pt}
\end{table}

\noindent\textbf{Performance on Adroit tasks.}
We also evaluate the performance of MOPP in Table \ref{tab:adroit} on more complex Adroit high-dimensional robotic manipulation tasks with sparse reward, involving twirling a pen, hammering a nail, opening a door and picking/ moving a ball. 
The Adroit datasets are particularly hard,
as the data are collected from a narrow expert data distributions (\texttt{expert}), human demonstrations (\texttt{human}), or a mixture of human demonstrations and imitation policies (\texttt{cloned}). 
Model-based offline RL methods are known to perform badly on such low-diversity datasets, as the dynamics models cannot be learned well (e.g., see results of MOPO). We compare MOPP with two more performant model-free offline RL algorithms, BCQ \cite{fujimoto2018off} and CQL \cite{kumar2020conservative}.
It is found that although MOPP is a model-based planning method, it performs surprisingly well in most of the cases. MOPP consistently outperforms the SOTA offline planning method MBOP, and in many tasks, it even outperforms the performant model-free offline RL baselines BCQ and CQL. The better performance of MOPP is a joint result of the inheritance of both an imitative behavior policy and more aggressive planning with the learned dynamics model.

\begin{figure*}[t]
	\centering
	\subfloat[Impacts of max-Q operation and trajectory pruning with different sampling aggressiveness ($\sigma_M$)]{
		\includegraphics[width=0.74\columnwidth]{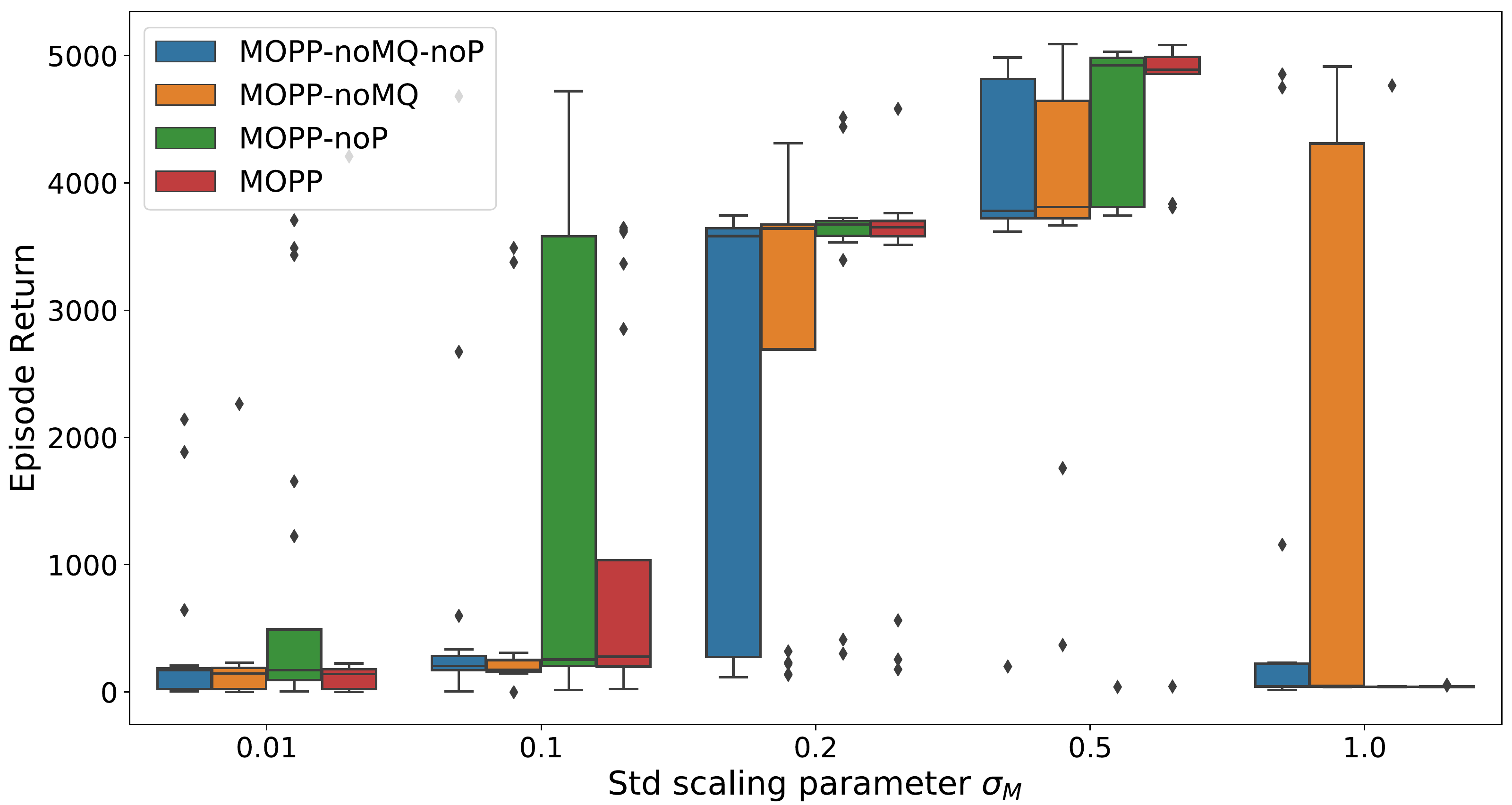}
	}\,\,
	\subfloat[Impacts of value function $V_b$ and max-Q operation with different planning horizon $H$]{
		\includegraphics[width=0.66\columnwidth]{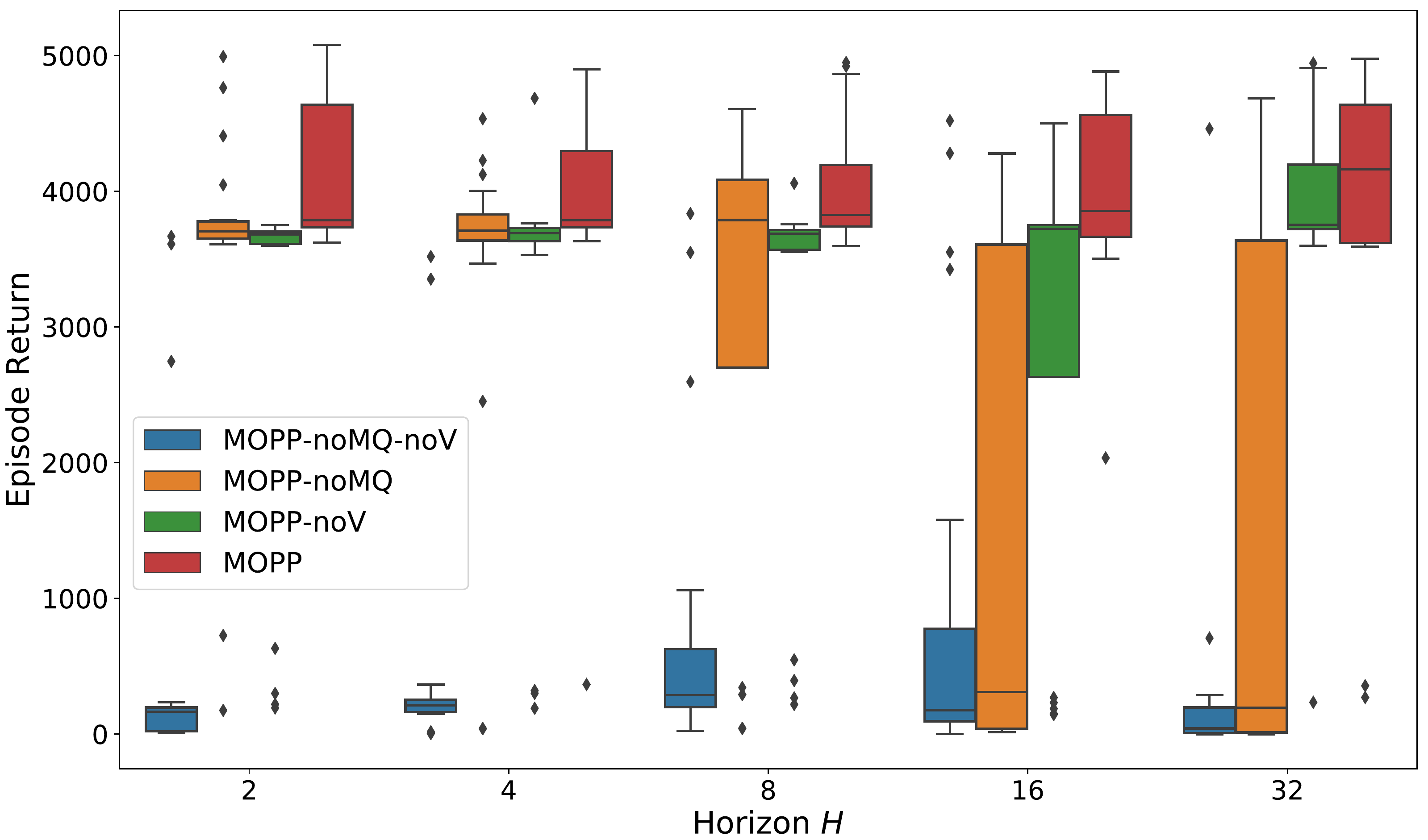}
	}\,\,
	\subfloat[Impact of uncertainty threshold $L$ ($\sigma_M=0.5$, $H=2$)]{
		\includegraphics[width= 0.53\columnwidth]{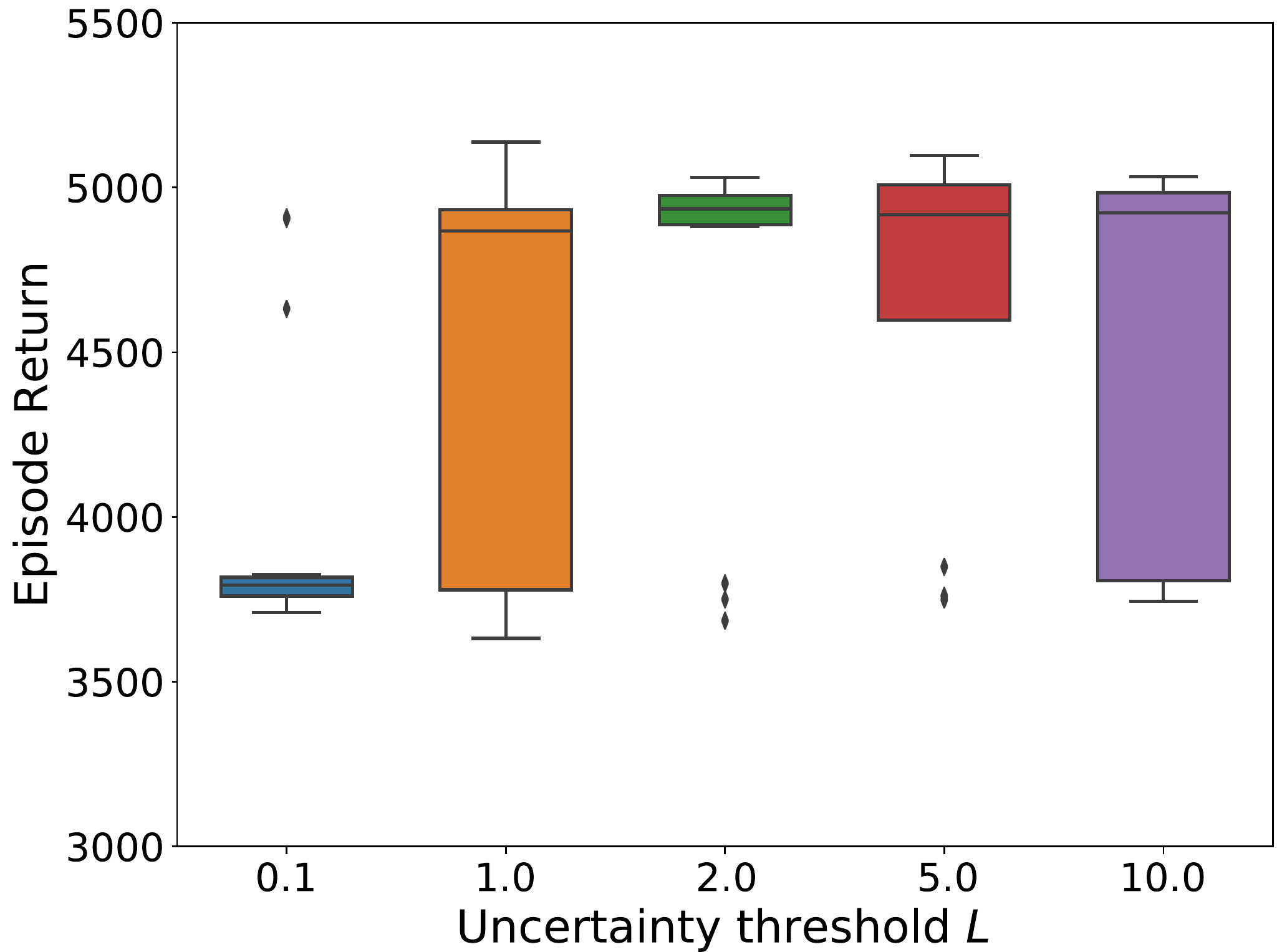}
	}
	\vspace{-6pt}
	\caption{Ablation study on the \texttt{walker2d-med-expert} task. \textbf{noMQ}, \textbf{noP}, \textbf{noV}  indicate MOPP without max-Q operation, trajectory pruning and value function $V_b$ respectively.}
	\label{fig:ablation}
	\vspace{-22pt}
\end{figure*}
\begin{figure*}[t]
	\centering
	\subfloat[Episode return under the new reward function]{
		\includegraphics[width=0.47\columnwidth]{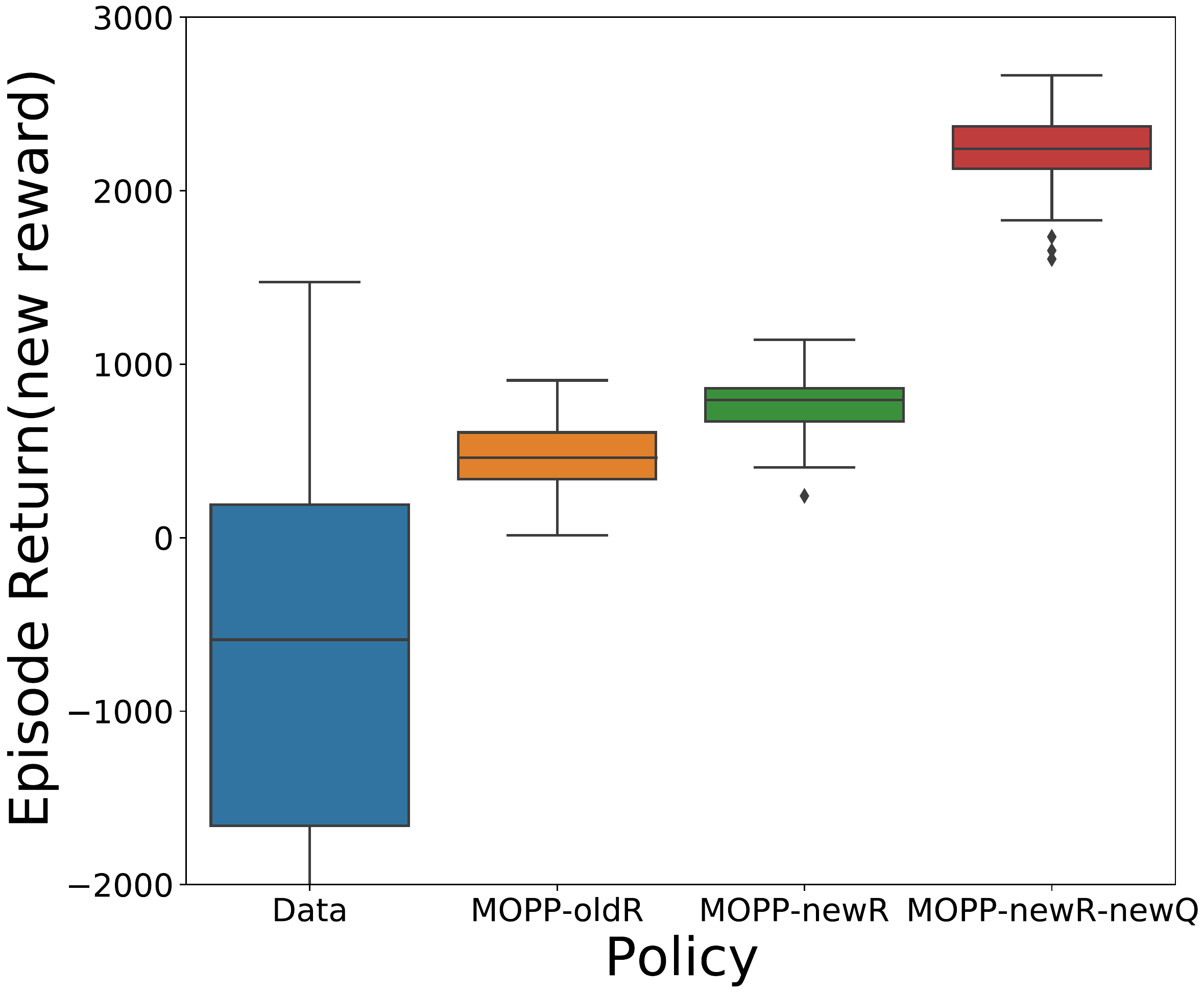}
	}\,\,
	\subfloat[The z-position of the optimized trajectories]{
		\includegraphics[width=0.47\columnwidth]{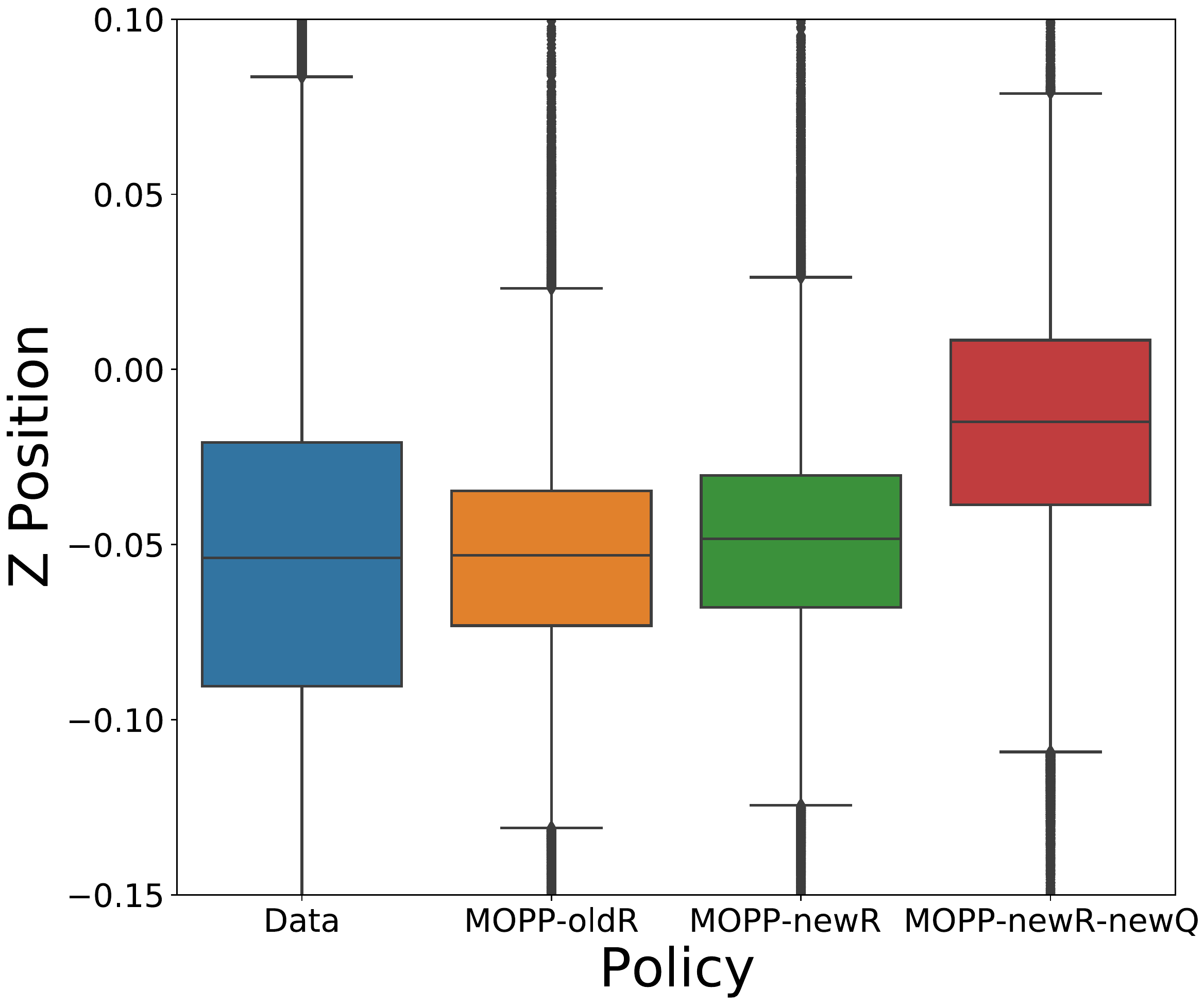}
	}\,\,
	\subfloat[Performance under x-velocity constraints.]{
		\includegraphics[width=0.76\columnwidth]{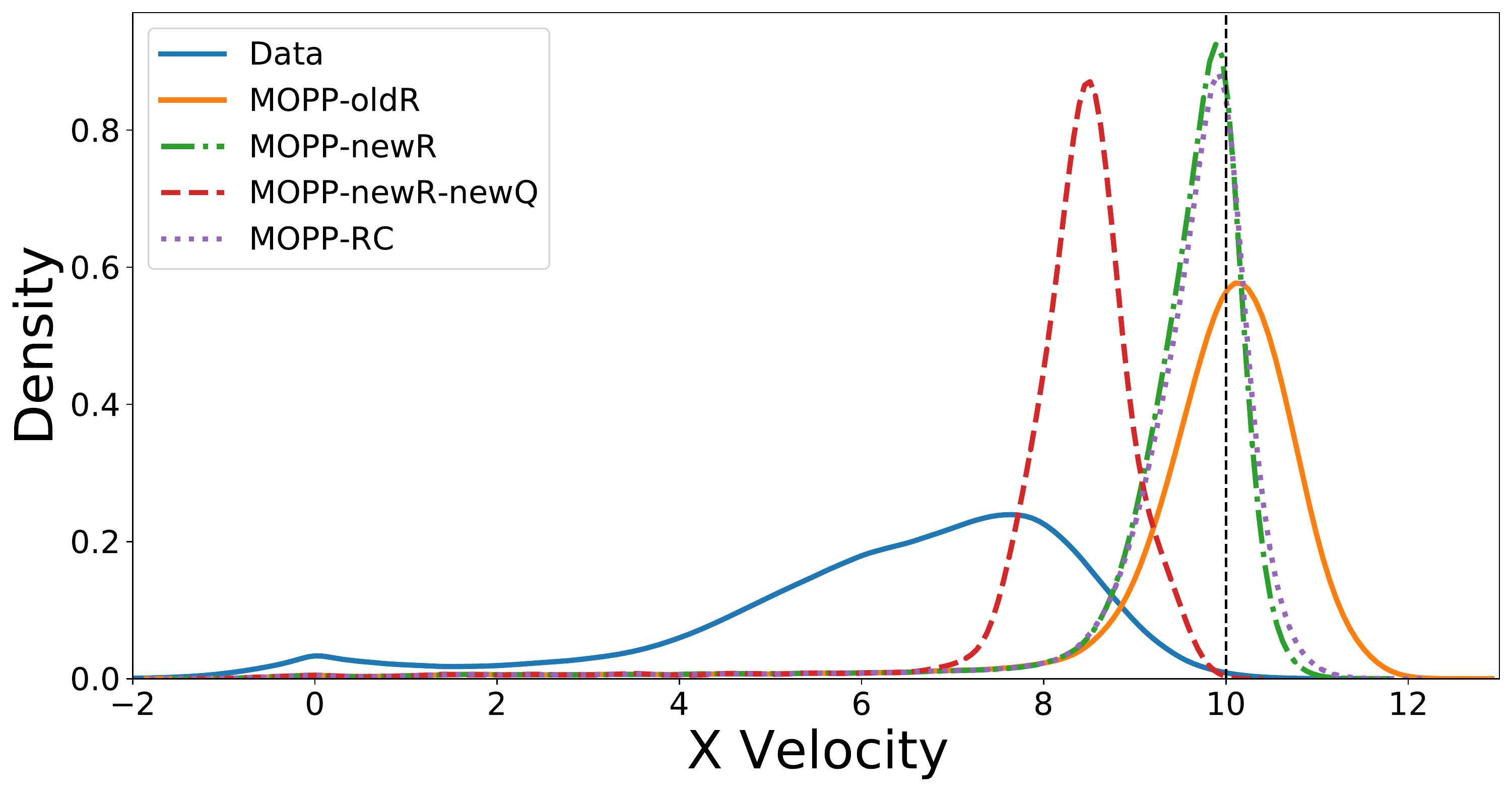}
	}
	
	\vspace{-8pt}
	\caption{Performance on \texttt{halfcheetah-jump} (a, b) and \texttt{halfcheetah-constrained} (c) tasks.
	}
	\label{fig:goal-conditioned-constrained}
	\vspace{-13pt}
\end{figure*}

\subsection{Ablation Study}
We conduct ablation experiments on \texttt{walker2d-med-expert} task to understand the impact of key elements in MOPP. 
We first investigate in Figure \ref{fig:ablation}(a) the level of sampling aggressiveness (controlled by std scaling parameter $\sigma_M$) on the performance of MOPP , as well as its relationship with the max-Q operation and trajectory pruning. It is observed that reasonably boosting the action sampling variance (e.g., increase $\sigma_M$ from $0.01$ to $0.5$) is beneficial. 
But overly aggressive exploration ($\sigma_M=1.0$) is detrimental, as it will introduce lots of undesired OOD samples during trajectory rollouts. When most trajectory rollouts are problematic, the trajectory pruning procedure is no longer effective, as there have to be at least $N_m$ trajectories in order to run the trajectory optimizer.
When $\sigma_M$ is not too large, trajectory pruning is effective to control the planning variance and produces better performance, as is shown in the difference between MOPP-noP and MOPP under $\sigma_M=0.5$. Moreover, the max-Q operation in the guided trajectory rollout increases the sampling aggressiveness. When $\sigma_M$ is moderate, MOPP achieves a higher score than MOPP-noMQ. But when $\sigma_M=1.0$, the less aggressive MOPP-noMQ is the only variant of MOPP that is still possible to produce high episode returns. These suggest that carefully choosing the degree of sampling aggressiveness is important for MOPP to achieve the best performance.

We further examine the impacts of value function $V_b$ and max-Q operation on different planning horizons in Figure \ref{fig:ablation}(b). It is observed that even with a very short horizon ($H=2$ and $4$), MOPP can attain good performance that is comparable to results using longer planning horizons. Moreover, MOPP achieves significantly higher scores compared with MOPP-noMQ-noV.
We found that using max-Q operation on sampled actions provides stronger improvements, as MOPP-noMQ consistently perform worse than MOPP-noV. 
This might because that max-Q operation is performed at every step, while the value function is only added to the end of the cumulative return of a trajectory, thus providing stronger guidance on trajectory rollouts towards potentially high reward actions.

Finally, Figure \ref{fig:ablation}(c) presents the impact of uncertainty threshold $L$ in trajectory pruning. We observe that both strictly avoid ($L=0.1$) or overly tolerant ($L=10.0$) unknown state-action pairs impact planning performance. Reasonably increase the tolerance of sample uncertainty  ($L=2.0$) to allow sufficient exploration leads to the best result with low variance. 

\vspace{-2pt}
\subsection{Evaluation on Control Flexibility}
A major advantage of planning methods lies in their flexibility to incorporate varying objectives and extra constraints.
These modifications can be easily incorporated in MOPP by revising the reward function or pruning out unqualified trajectory rollouts during operation. We construct two tasks to evaluated the control flexibility of MOPP:
\begin{itemize}[leftmargin=*,topsep=0pt,noitemsep]
	\item \texttt{halfcheetah-jump}: This task adds incentives on the z-position in the original reward function of halfcheetah, encouraging agent to run while jumping as high as possible.
	
	\item \texttt{halfcheetah-constrained} task adds a new constraint (x-velocity$\leq$10) to restrain agent from having very high x-velocity. Two ways are used to incorporate the constraint: 1) add reward penalty for x-velocity$>$10; 2) add penalties on the uncertainty measures $\mU$ to allow trajectory pruning to filter out constraint violating trajectory rollouts.
\end{itemize}

Figure \ref{fig:goal-conditioned-constrained}(a), (b) shows the performance of MOPP on the \texttt{halfcheetah-jump} task. By simply changing to the new reward function (MOPP-newR), MOPP is able to adapt and improve upon the average performance level in data and the original model (MOPP-oldR). The performance will be further improved by re-evaluating the Q-function (MOPP-newR-newQ). 
The offline evaluated value function $Q_b$ and the max-Q operation
could have negative impact when the reward function is drastically different. In such cases, one only needs to re-evaluate a sub-component ($Q_b$ under the new reward) of MOPP to guarantee the best performance rather than re-train the whole model as in typical RL settings. 
Evaluating $Q_b$ via FQE is achieved by simple supervised learning, which is computationally very cheap compared with a costly RL procedure (see Appendix for discussion on computation performance).



Figure \ref{fig:goal-conditioned-constrained}(c) presents the performance on the \texttt{halfcheetah} \texttt{-constrained} task. The original MOPP without constraint (MOPP-oldR) has lots of constraint violations (x-velocity$>$10). Incorporating a constraint penalty in reward (MOPP-newR) and pruning out constraint violating trajectories (MOPP-RC) achieve very similar performance. Both models effectively reduce constraint violations and have limited performance deterioration due to the extra constraint. Adding constraint penalty in the reward function while 
re-evaluating the $Q_b$ via FQE (MOPP-newR-newQ) leads to the safest policy. 

\vspace{-2pt}
\section{Conclusion}
We propose MOPP, a light-weighted model-based offline planning algorithm for real-world control tasks when online training is forbidden.
MOPP is built upon an MPC framework that leverages behavior policies and dynamics models learned from an offline dataset to perform planning. 
MOPP avoids over-restrictive planning while 
enabling offline learning by encouraging more aggressive trajectory rollout guided by the learned behavior policy, and prunes out problematic trajectories by evaluating the uncertainty of dynamics models.
Although MOPP is a planning method, benchmark experiments show that 
it provides competitive performance compared with the state-of-the-art offline RL and model-based planning methods. 


\newpage
\small
\bibliographystyle{named}
\bibliography{ijcai22}

\begin{thebibliography}{}

\bibitem[\protect\citeauthoryear{Argenson and
  Dulac-Arnold}{2021}]{argenson2020model}
Arthur Argenson and Gabriel Dulac-Arnold.
\newblock Model-based offline planning.
\newblock In {\em International Conference on Learning Representations}, 2021.

\bibitem[\protect\citeauthoryear{Botev \bgroup \em et al.\egroup
  }{2013}]{botev2013cross}
Zdravko~I Botev, Dirk~P Kroese, Reuven~Y Rubinstein, and Pierre L’Ecuyer.
\newblock The cross-entropy method for optimization.
\newblock In {\em Handbook of statistics}, volume~31, pages 35--59. Elsevier,
  2013.

\bibitem[\protect\citeauthoryear{Buckman \bgroup \em et al.\egroup
  }{2020}]{buckman2020importance}
Jacob Buckman, Carles Gelada, and Marc~G Bellemare.
\newblock The importance of pessimism in fixed-dataset policy optimization.
\newblock In {\em International Conference on Learning Representations}, 2020.

\bibitem[\protect\citeauthoryear{Chua \bgroup \em et al.\egroup
  }{2018}]{chua2018deep}
Kurtland Chua, Roberto Calandra, Rowan McAllister, and Sergey Levine.
\newblock Deep reinforcement learning in a handful of trials using
  probabilistic dynamics models.
\newblock In {\em Advances in Neural Information Processing Systems}, pages
  4754--4765, 2018.

\bibitem[\protect\citeauthoryear{Fu \bgroup \em et al.\egroup
  }{2020}]{fu2020d4rl}
Justin Fu, Aviral Kumar, Ofir Nachum, George Tucker, and Sergey Levine.
\newblock D4rl: Datasets for deep data-driven reinforcement learning.
\newblock {\em arXiv preprint arXiv:2004.07219}, 2020.

\bibitem[\protect\citeauthoryear{Fujimoto and
  Gu}{2021}]{fujimoto2021minimalist}
Scott Fujimoto and Shixiang~Shane Gu.
\newblock A minimalist approach to offline reinforcement learning.
\newblock In {\em Advances in Neural Information Processing Systems}, 2021.

\bibitem[\protect\citeauthoryear{Fujimoto \bgroup \em et al.\egroup
  }{2019}]{fujimoto2018off}
Scott Fujimoto, David Meger, and Doina Precup.
\newblock Off-policy deep reinforcement learning without exploration.
\newblock In {\em International Conference on Machine Learning}, pages
  2052--2062. PMLR, 2019.

\bibitem[\protect\citeauthoryear{Germain \bgroup \em et al.\egroup
  }{2015}]{germain2015made}
Mathieu Germain, Karol Gregor, Iain Murray, and Hugo Larochelle.
\newblock Made: Masked autoencoder for distribution estimation.
\newblock In {\em International Conference on Machine Learning}, pages
  881--889, 2015.

\bibitem[\protect\citeauthoryear{Ghasemipour \bgroup \em et al.\egroup
  }{2021}]{ghasemipour2020emaq}
Seyed Kamyar~Seyed Ghasemipour, Dale Schuurmans, and Shixiang~Shane Gu.
\newblock Emaq: Expected-max q-learning operator for simple yet effective
  offline and online rl.
\newblock In {\em International Conference on Machine Learning}, pages
  3682--3691. PMLR, 2021.

\bibitem[\protect\citeauthoryear{Haarnoja \bgroup \em et al.\egroup
  }{2018}]{haarnoja2018soft}
Tuomas Haarnoja, Aurick Zhou, Pieter Abbeel, and Sergey Levine.
\newblock Soft actor-critic: Off-policy maximum entropy deep reinforcement
  learning with a stochastic actor.
\newblock In {\em International Conference on Machine Learning}, pages
  1861--1870. PMLR, 2018.

\bibitem[\protect\citeauthoryear{Janner \bgroup \em et al.\egroup
  }{2019}]{janner2019trust}
Michael Janner, Justin Fu, Marvin Zhang, and Sergey Levine.
\newblock When to trust your model: Model-based policy optimization.
\newblock In {\em Advances in Neural Information Processing Systems}, pages
  12519--12530, 2019.

\bibitem[\protect\citeauthoryear{Kidambi \bgroup \em et al.\egroup
  }{2020}]{kidambi2020morel}
Rahul Kidambi, Aravind Rajeswaran, Praneeth Netrapalli, and Thorsten Joachims.
\newblock Morel: Model-based offline reinforcement learning.
\newblock In {\em Advances in Neural Information Processing Systems}, pages
  21810--21823, 2020.

\bibitem[\protect\citeauthoryear{Kostrikov \bgroup \em et al.\egroup
  }{2021}]{kostrikov2021offline}
Ilya Kostrikov, Rob Fergus, Jonathan Tompson, and Ofir Nachum.
\newblock Offline reinforcement learning with fisher divergence critic
  regularization.
\newblock In {\em International Conference on Machine Learning}, 2021.

\bibitem[\protect\citeauthoryear{Kumar \bgroup \em et al.\egroup
  }{2019}]{kumar2019stabilizing}
Aviral Kumar, Justin Fu, Matthew Soh, George Tucker, and Sergey Levine.
\newblock Stabilizing off-policy q-learning via bootstrapping error reduction.
\newblock In {\em Advances in Neural Information Processing Systems}, pages
  11761--11771, 2019.

\bibitem[\protect\citeauthoryear{Kumar \bgroup \em et al.\egroup
  }{2020}]{kumar2020conservative}
Aviral Kumar, Aurick Zhou, George Tucker, and Sergey Levine.
\newblock Conservative q-learning for offline reinforcement learning.
\newblock In {\em Advances in Neural Information Processing Systems}, pages
  1179--1191, 2020.

\bibitem[\protect\citeauthoryear{Le \bgroup \em et al.\egroup
  }{2019}]{le2019batch}
Hoang~M Le, Cameron Voloshin, and Yisong Yue.
\newblock Batch policy learning under constraints.
\newblock In {\em International Conference on Machine Learning}, pages
  3703--3712. PMLR, 2019.

\bibitem[\protect\citeauthoryear{Levine \bgroup \em et al.\egroup
  }{2016}]{levine2016end}
Sergey Levine, Chelsea Finn, Trevor Darrell, and Pieter Abbeel.
\newblock End-to-end training of deep visuomotor policies.
\newblock {\em The Journal of Machine Learning Research}, 17(1):1334--1373,
  2016.

\bibitem[\protect\citeauthoryear{Liu \bgroup \em et al.\egroup
  }{2020}]{liu2020provably}
Yao Liu, Adith Swaminathan, Alekh Agarwal, and Emma Brunskill.
\newblock Provably good batch off-policy reinforcement learning without great
  exploration.
\newblock In {\em Advances in Neural Information Processing Systems}, pages
  1264--1274, 2020.

\bibitem[\protect\citeauthoryear{Lowrey \bgroup \em et al.\egroup
  }{2019}]{lowrey2018plan}
Kendall Lowrey, Aravind Rajeswaran, Sham Kakade, Emanuel Todorov, and Igor
  Mordatch.
\newblock Plan online, learn offline: Efficient learning and exploration via
  model-based control.
\newblock In {\em International Conference on Learning Representations}, 2019.

\bibitem[\protect\citeauthoryear{Nagabandi \bgroup \em et al.\egroup
  }{2020}]{nagabandi2020deep}
Anusha Nagabandi, Kurt Konolige, Sergey Levine, and Vikash Kumar.
\newblock Deep dynamics models for learning dexterous manipulation.
\newblock In {\em Conference on Robot Learning}, pages 1101--1112. PMLR, 2020.

\bibitem[\protect\citeauthoryear{Rajeswaran \bgroup \em et al.\egroup
  }{2018}]{Rajeswaran-RSS-18}
Aravind Rajeswaran, Vikash Kumar, Abhishek Gupta, Giulia Vezzani, John
  Schulman, Emanuel Todorov, and Sergey Levine.
\newblock Learning complex dexterous manipulation with deep reinforcement
  learning and demonstrations.
\newblock In {\em Proceedings of Robotics: Science and Systems}, June 2018.

\bibitem[\protect\citeauthoryear{Silver \bgroup \em et al.\egroup
  }{2017}]{silver2017mastering}
David Silver, Julian Schrittwieser, Karen Simonyan, Ioannis Antonoglou, Aja
  Huang, Arthur Guez, Thomas Hubert, Lucas Baker, Matthew Lai, Adrian Bolton,
  et~al.
\newblock Mastering the game of go without human knowledge.
\newblock {\em nature}, 550(7676):354--359, 2017.

\bibitem[\protect\citeauthoryear{Wang and Ba}{2020}]{wang2019exploring}
Tingwu Wang and Jimmy Ba.
\newblock Exploring model-based planning with policy networks.
\newblock In {\em International Conference on Learning Representations}, 2020.

\bibitem[\protect\citeauthoryear{Williams \bgroup \em et al.\egroup
  }{2017}]{williams2017model}
Grady Williams, Andrew Aldrich, and Evangelos~A Theodorou.
\newblock Model predictive path integral control: From theory to parallel
  computation.
\newblock {\em Journal of Guidance, Control, and Dynamics}, 40(2):344--357,
  2017.

\bibitem[\protect\citeauthoryear{Wu \bgroup \em et al.\egroup
  }{2019}]{wu2019behavior}
Yifan Wu, George Tucker, and Ofir Nachum.
\newblock Behavior regularized offline reinforcement learning.
\newblock {\em arXiv preprint arXiv:1911.11361}, 2019.

\bibitem[\protect\citeauthoryear{Xu \bgroup \em et al.\egroup
  }{2021}]{xu2021offline}
Haoran Xu, Xianyuan Zhan, Jianxiong Li, and Honglei Yin.
\newblock Offline reinforcement learning with soft behavior regularization.
\newblock {\em arXiv preprint arXiv:2110.07395}, 2021.

\bibitem[\protect\citeauthoryear{Xu \bgroup \em et al.\egroup
  }{2022}]{xu2021constraints}
Haoran Xu, Xianyuan Zhan, and Xiangyu Zhu.
\newblock Constraints penalized q-learning for safe offline reinforcement
  learning.
\newblock In {\em Proceedings of the AAAI Conference on Artificial
  Intelligence}, 2022.

\bibitem[\protect\citeauthoryear{Yu \bgroup \em et al.\egroup
  }{2020}]{yu2020mopo}
Tianhe Yu, Garrett Thomas, Lantao Yu, Stefano Ermon, James Zou, Sergey Levine,
  Chelsea Finn, and Tengyu Ma.
\newblock Mopo: Model-based offline policy optimization.
\newblock In {\em Advances in Neural Information Processing Systems}, pages
  14129--14142, 2020.

\bibitem[\protect\citeauthoryear{Zhan \bgroup \em et al.\egroup
  }{2022}]{zhan2021deepthermal}
Xianyuan Zhan, Haoran Xu, Yue Zhang, Yusen Huo, Xiangyu Zhu, Honglei Yin, and
  Yu~Zheng.
\newblock Deepthermal: Combustion optimization for thermal power generating
  units using offline reinforcement learning.
\newblock In {\em Proceedings of the AAAI Conference on Artificial
  Intelligence}, 2022.

\end{thebibliography}

\normalsize
\clearpage
\appendix
\section*{Appendix}
\section{Detailed Experiment Settings}
\subsection{Benchmark Datasets}
\begin{figure*}[t]
	\centering
	\includegraphics[width=\textwidth]{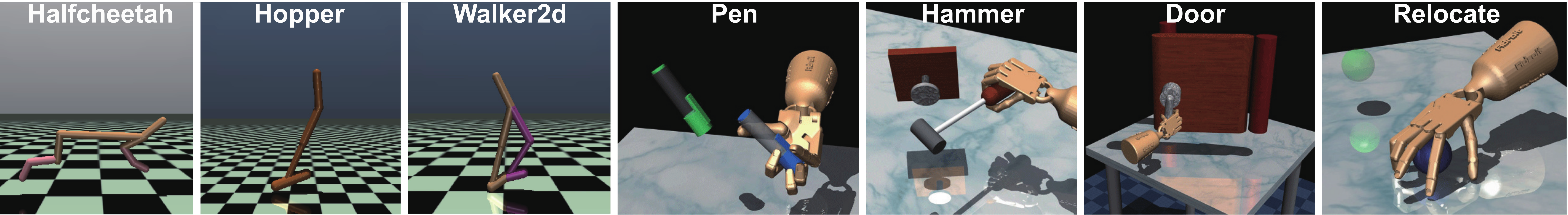}
	\caption{Visualization of the evaluated tasks}
	\label{fig:tasks}
	
	\includegraphics[width=1.75\columnwidth]{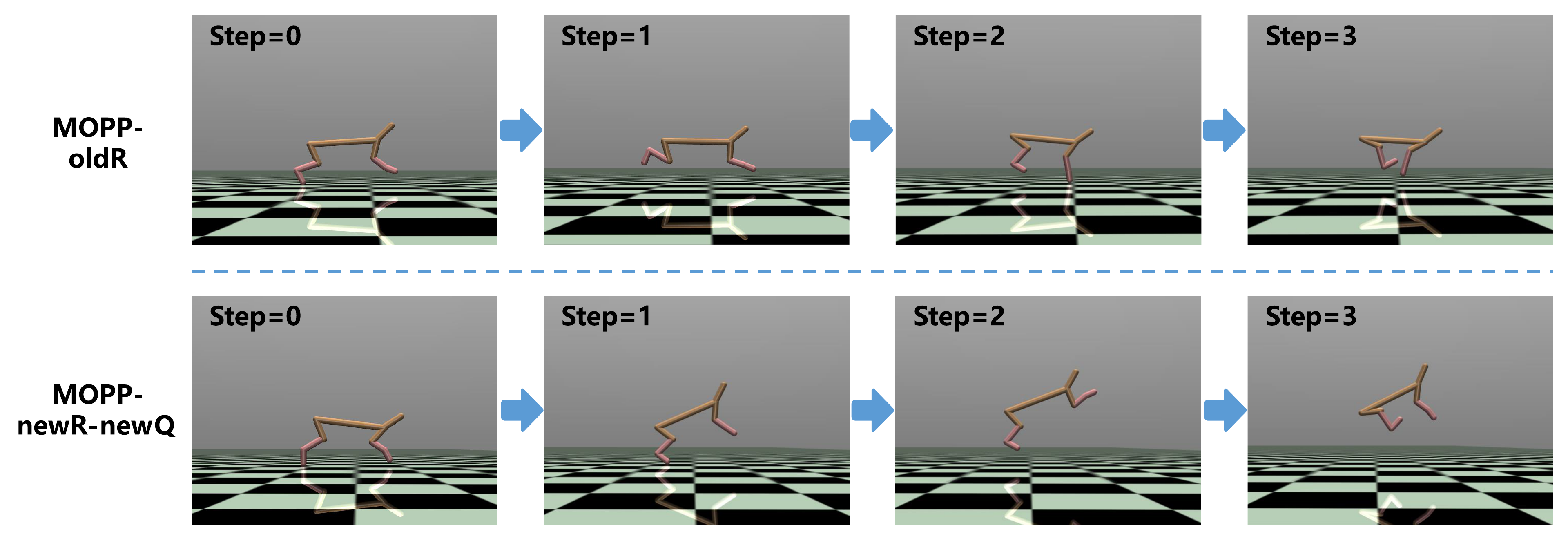}
	\caption{
		Illustrations of the original \texttt{halfcheetah} and the \texttt{halfcheetah-jump} tasks. The top row shows the results of MOPP with the original objective (MOPP-oldR). The bottom row shows the results obtained using MOPP with the new reward function and re-evaluated Q-function (MOPP-newR-newQ). It is observed that the halfcheetah agent using MOPP-newR-newQ adapts to the new objective that is running while jumping as high as possible.
	}
	\label{fig:visualize-cheetah-jump}
\end{figure*}

We evaluate the performance of MOPP on the popular D4RL MuJoCo tasks (Halfcheetah, Hopper, Walker2d) and the more complex Adroit hand manipulation tasks (Pen, Hammer, Door, Relocate) \cite{fu2020d4rl}. These tasks are visualized in Figure \ref{fig:tasks}.

\subsubsection{MoJoCo tasks}
We test 12 problem settings in the D4RL MoJoCo benchmark, including three environments: \texttt{halfcheetah}, \texttt{hopper}, \texttt{walker2d} and four dataset types: \texttt{random}, \texttt{medium}, \texttt{mixed} and \texttt{med-expert}.	
The MoJoCo benchmark datasets are generated as follows:
\begin{itemize}[leftmargin=*]
	\item \textbf{random}: generated using a random policy to roll out 1M steps.
	\item \textbf{medium}: generated using a partially trained SAC policy \cite{haarnoja2018soft} to roll out 1M steps.
	\item \textbf{mixed}: train an SAC policy until reaching a predefined threshold, and take the replay buffer as the dataset. This dataset is also termed \texttt{medium-replay} in the latest version of the D4RL paper.
	\item \textbf{med-expert}: generated by combining 1M samples from a expert policy and 1M samples from a partially trained policy.
\end{itemize}


\subsubsection{Adroit tasks}
The Adroit hand manipulation environment \cite{Rajeswaran-RSS-18} involves controlling a 24-DoF simulated Shadow Hand robot with twirling a pen (Pen), hammering a nail (Hammer), opening a door (Door) and picking up and moving a ball (Relocate). Adroit tasks are substantially more challenging than the MoJoCo tasks, as the dataset is collected from human demonstrations and fine-tuned online RL expert policies with narrow data distributions on sparse reward, high-dimensional control tasks. The Adroit tasks have three dataset types:
\begin{itemize}[leftmargin=*]
	\item \textbf{human}: contains a small amount of demonstration data from a human, 25 trajectories per task.
	\item \textbf{expert}: contains a larger amount of expert data from a fine-tuned online RL policy.
	\item \textbf{cloned}: generated by training an imitation policy on the demonstrations, running the policy, and mixing data at a 50-50 ratio with demonstrations.
\end{itemize}

\subsection{Hyperparameters Used in the D4RL Benchmark Experiments}
In Table \ref{tab:h-param}, we present the hyperparameters used for runs of MOPP on the D4RL benchmark. We kept most hyperparameter settings close to MBOP \cite{argenson2020model} to make our results comparable to those reported in the MBOP paper. In the ablation study, all the hyperparameters of MOPP are the same as that in Table \ref{tab:h-param} except the varied parameters in the ablation experiments. The hyperparameter $L$ is selected based on the 85th percentile value of the uncertainty measures computed from the offline dataset.
In addition to the hyperparameters reported in the table, for all experiments, we use $N_m = 0.2 \lfloor N \rfloor$, $K_1=K_2=3$ and $K_Q=10$ (see Algorithm 1 in the main article for details).

\begin{table}[H]
	\centering
	\small
	\begin{tabular}{l|c|c|c|c|c|c}
		\toprule
		\multicolumn{7}{c}{MoJoCo HalfCheetah} \\
		\midrule
		Dataset & $H$  & $\kappa$ & $\beta$ & $L$& $\sigma_M$ & $N$ \\
		\midrule
		random & 4& 3 & 0 & 4& 1.15& 100   \\ 
		medium & 2& 3 & 0& 5& 0.45 & 100   \\ 
		mixed & 4& 3 & 0& 5& 0.5 & 100  \\ 
		med-expert & 2& 1 & 0& 7& 0.55 & 100   \\ 
		\midrule
		\midrule
		\multicolumn{7}{c}{MoJoCo Hopper} \\
		\midrule
		Dataset & $H$  & $\kappa$ & $\beta$ & $L$& $\sigma_M$ &  $N$ \\
		\midrule
		random & 4& 10 & 0& 0.5& 0.65& 100    \\ 
		medium & 4& 0.3 & 0& 1& 0.25 & 100    \\ 
		mixed & 4& 0.3 & 0& 1& 0.6 & 100   \\ 
		med-expert & 10& 3 & 0& 1& 0.4 & 100   \\ 
		\midrule
		\midrule
		\multicolumn{7}{c}{MoJoCo Walker2d} \\
		\midrule
		Dataset & $H$  & $\kappa$ & $\beta$ & $L$& $\sigma_M$ & $N$ \\
		\midrule
		random & 8& 0.3 & 0& 8& 0.05& 1000     \\ 
		medium & 2& 0.1& 0& 7& 0.55& 1000    \\ 
		mixed & 8& 3& 0& 8& 0.2& 1000    \\ 
		med-expert & 2& 1& 0& 7& 0.4& 1000    \\ 
		\midrule
		\midrule
		\multicolumn{7}{c}{Adroit Pen} \\
		\midrule
		Dataset & $H$  & $\kappa$ & $\beta$ & $L$& $\sigma_M$ & $N$ \\
		\midrule
		human & 4 & 0.3 & 0 & 0.1 & 0.8 & 100     \\ 
		cloned & 4 & 0.3 & 0 & 1.7 & 0.8 & 100    \\ 
		expert & 4 & 0.03 & 0 & 4.4 & 0.8 & 100    \\
		\midrule
		\midrule
		\multicolumn{7}{c}{Adroit Hammer} \\
		\midrule
		Dataset & $H$  & $\kappa$ & $\beta$ & $L$& $\sigma_M$ & $N$ \\
		\midrule
		human & 4 & 0.3 & 0 & 0.3 & 1.0 & 100     \\ 
		cloned & 4 & 0.3 & 0 & 0.5 & 0.8 & 100    \\ 
		expert & 4 & 0.3 & 0 & 1.4 & 0.7 & 100    \\
		\midrule
		\midrule
		\multicolumn{7}{c}{Adroit Door} \\
		\midrule
		Dataset & $H$  & $\kappa$ & $\beta$ & $L$& $\sigma_M$ & $N$ \\
		\midrule
		human & 4 & 0.3 & 0 & 1.2 & 0.8 & 100     \\ 
		cloned & 4 & 0.3 & 0 & 0.3 & 0.8 & 100    \\ 
		expert & 4 & 0.03 & 0 & 0.1 & 0.7 & 100    \\
		\midrule
		\midrule
		\multicolumn{7}{c}{Adroit Relocate} \\
		\midrule
		Dataset & $H$  & $\kappa$ & $\beta$ & $L$& $\sigma_M$ & $N$ \\
		\midrule
		human & 4 & 0.3 & 0 & 1.0 & 0.8 & 100     \\ 
		cloned & 4 & 0.3 & 0 & 0.4 & 0.8 & 100    \\ 
		expert & 16 & 0.3 & 0 & 0.1 & 0.4 & 100    \\
		\bottomrule
	\end{tabular} 
	\caption{Hyperparameters of MOPP used in the D4RL benchmark experiments}
	\label{tab:h-param}
\end{table}

\subsection{Flexibility of Incorporating Varying Objectives and Constraints}
We modify the original \texttt{halfcheetah} task and construct two new tasks (\texttt{halfcheetah-jump} and \texttt{halfcheetah} \texttt{-constrained}) to evaluate the flexibility and generalizibility of MOPP on new tasks with varying objectives and extra constraints. In both tasks, MOPP is trained using the entire 1M steps training replay buffer of SAC on the original \texttt{halfcheetah} task. We modify the reward function or introduce rollout constraints in MOPP during planning. To test for best performance and examine the impact of max-Q operation, we also report the results of MOPP with re-evaluated Q-functions under the new reward function via FQE.
The results of MOPP and its variants of the new tasks are reported in Figure \ref{fig:goal-conditioned-constrained} in the main article. All results are computed based on 6 random seeds, with 20 episode runs per seed.

\subsubsection{Control under varying objective}
In the \texttt{halfcheetah} \texttt{-jump} task, we modify the objective of the original halfcheetah agent, which encourages the agent to have higher z-position, leading to a run and jump behavior. The modified reward function in the new objective is:
\begin{equation}
\label{eq:goal-conditioned}
r' = \alpha_r \cdot r + (1-\alpha_r)\cdot 100 \cdot z
\end{equation}
where $r$ is the original reward of the \texttt{halfcheetah} task, and $z$ denotes the z-position of the halfcheetah agent. In our experiment, $\alpha_r$ is set as 0.4. Note that our \texttt{halfcheetah-jump} task is different from the one reported in the MOPO paper \cite{yu2020mopo} which sets the maximum velocity to be 3 in both behavior policy and its revised reward to only encourage the jump behavior.


\subsubsection{Constrained control}
In the  \texttt{halfcheetah-} \texttt{constrained} task, we introduce a state-based constraint to the original \texttt{halfcheetah} task. We constrain the velocity along the x-axis ($v_x$) of the halfcheetah agent below a certain threshold (10 m/s).
Two implementations are tested in our experiments:

\begin{itemize}[leftmargin=*]
	\item \textbf{Reward penalization:} Adding a reward penalty for $v_x>10$ in the reward function:
	\begin{equation}
	\label{eq:r-constrained-control}
	r' = \alpha_c \cdot r + (1-\alpha_c) \cdot 100 \cdot \min (10-v_x,0)
	\end{equation}
	where $r$ is the original reward of the \texttt{halfcheetah} task and the weight $\alpha_c$ is set as 0.5 in the experiments.
	
	\item \textbf{Rollout constraint:} We filter out the trajectories that violate the state-based constraint during trajectory pruning in MOPP. This is achieved as a rollout constraint by adding penalties to the uncertainty measures $U_{n,t}$ of the constraint violating trajectory rollouts.
	\begin{equation}
	\label{eq:rollout-constrained-control}
	U_{n,t}' = U_{n,t} + 100 \cdot \max(v_x-10,0)
	\end{equation}
	Note that due to the existence of the minimum number of required trajectories $N_m$ to run the trajectory optimizer, it is possible that some unsafe trajectories will remain after trajectory pruning if most of the trajectory rollouts violate the constraint. The advantage of rollout constraint is that it does not alter the reward function, thus has no impact on the Q-function learned from the behavioral data.
\end{itemize}

\begin{table*}[t]
	\centering
	\scriptsize
	\begin{tabular}{l|l|c|c|c|c|c}	
		\toprule
		\textbf{Dataset type}    & \textbf{Environment}  & \textbf{MBOP (MBOP $f_b'$)} & \textbf{MOPP-ADM $f_m$ (BC)}  & \textbf{MOPP-ADM $f_m$ (ADM $f_b$)} & \textbf{MOPP-FC $f_m$ (BC)}  & \textbf{MOPP-FC $f_m$ (ADM $f_b$)}  \\ 
		\midrule
		random & halfcheetah & 6.3$\pm$4.0 (0.0$\pm$0.0)& 9.1$\pm$0.2 (2.2$\pm$2.2) &\textbf{9.4$\pm$2.6} (2.2$\pm$2.2)  & 7.9 $\pm$ 0.3 (2.2$\pm$2.2) & 7.9$\pm$0.3 (2.2$\pm$2.2)      \\ 
		random & hopper & 10.8$\pm$0.3 (9.0$\pm$0.2) & 11.9$\pm$0.1 (10.0$\pm$0.7) &\textbf{13.7$\pm$2.5} (9.8$\pm$0.7) & 11.8$\pm$0.2 (10.0$\pm$0.7) & 12.1$\pm$0.2 (9.8$\pm$0.7)     \\ 
		random & walker2d & \textbf{8.1$\pm$5.5} (0.1$\pm$0.0) &4.9$\pm$0.9 (6.2$\pm$2.2) & 6.3$\pm$0.1 (2.6$\pm$0.1) & 6.7$\pm$0.2 (6.2$\pm$2.2) & 6.5$\pm$0.1 (2.6$\pm$0.1)      \\ 
		\midrule
		medium & halfcheetah & 44.6$\pm$0.8 (35.0$\pm$2.5) & 44.5$\pm$0.3 (36.7$\pm$4.1) &44.7$\pm$2.6 (36.6$\pm$4.7) & 45.0$\pm$0.4(36.7$\pm$4.1) & \textbf{45.1$\pm$0.3} (36.6$\pm$4.7)    \\ 
		medium & hopper &\textbf{48.8$\pm$26.8} (48.1$\pm$26.2) & 28.2$\pm$8.8 (30.4$\pm$0.9) &31.8$\pm$1.3 (30.0$\pm$0.8) & 28.0$\pm$9.1 (30.4$\pm$0.9) & 27.3$\pm$9.9 (30.0$\pm$0.8)      \\ 
		medium & walker2d &41.0$\pm$29.4 (15.4$\pm$24.7) & \textbf{82.3$\pm$0.9} (15.0$\pm$19.8) &80.7$\pm$1.0 (15.6$\pm$22.5) & 81.3$\pm$4.2 (15.0$\pm$19.8) & 81.4$\pm$1.3 (15.6$\pm$22.5)      \\
		\midrule
		mixed & halfcheetah &42.3$\pm$0.9 (0.0$\pm$0.0) & 41.4$\pm$1.9 (31.8$\pm$7.2) &\textbf{43.1$\pm$4.3} (32.7$\pm$7.7) & 40.8$\pm$0.8 (31.8$\pm$7.2) & 40.3$\pm$0.7 (32.7$\pm$7.7)     \\ 
		mixed & hopper &12.4$\pm$5.8 (9.5$\pm$6.9) & 30.6$\pm$2.7 (20.0$\pm$7.7) &\textbf{32.3$\pm$5.9} (28.2$\pm$4.3) & 30.5$\pm$2.9 (20.0$\pm$7.7) & 28.0$\pm$0.9 (28.2$\pm$4.3)      \\ 
		mixed & walker2d & 9.7$\pm$5.3 (11.5$\pm$7.3) & 16.5$\pm$7.4 (12.9$\pm$4.5) &\textbf{18.5$\pm$8.4} (12.9$\pm$5.7) & 15.4$\pm$7.4 (12.9$\pm$4.5) & 15.0$\pm$6.7 (12.9$\pm$5.7)   \\
		\midrule
		med-expert & halfcheetah &105.9$\pm$17.8 (90.8$\pm$26.9) & 103.7$\pm$11.0 (37.6$\pm$6.5) &\textbf{106.2$\pm$5.1} (37.6$\pm$6.5)  &  96.7$\pm$10.1 (37.6$\pm$6.5) & 86.9$\pm$22.1 (37.6$\pm$6.5)     \\  
		med-expert & hopper &55.1$\pm$44.3 (15$\pm$8.7) & 94.4$\pm$31.6 (34.1$\pm$18.7) &95.4$\pm$28.0 (44.3$\pm$28.4)  & \textbf{103.9$\pm$23.8} (34.1$\pm$18.7) & 51.6$\pm$36.0 (44.3$\pm$28.4)     \\ 
		med-expert & walker2d &70.2$\pm$36.2 (65.5$\pm$40.2)& 88.3$\pm$38.8 (6.6$\pm$13.8) &\textbf{92.9$\pm$14.1} (13.5$\pm$24.2)  & 87.9$\pm$38.8 (6.6$\pm$13.8) & 85.6$\pm$10.6 (13.5$\pm$24.2)   \\ 
		\bottomrule
	\end{tabular}
	
	\caption{Ablation results on  D4RL MuJoCo tasks for MOPP on using feed-forward (\textbf{FC} $f_m$) and ADM (\textbf{ADM} $f_m$) dynamics model as well as BC and ADM behavior policy $f_b$. The feed-forward dynamics model (\textbf{FC} $f_m$) is the same as that reported in MBOP \protect\cite{argenson2020model}. The scores are normalized between 0 to 100 (0 and 100 correspond to a random policy and an expert SAC policy respectively). The scores of the used behavior policies (\textbf{MBOP} $f_b'$, \textbf{BC} and \textbf{ADM} $f_b$) in the parentheses. All results are computed based on 5 random seeds, with 20 episode runs per seed.
	}
	\label{tab:adm_ablation}
	
\end{table*}

\begin{figure}[t]
	\centering
	\subfloat[Episode return under the original reward function]{
		\includegraphics[width=0.48\columnwidth]{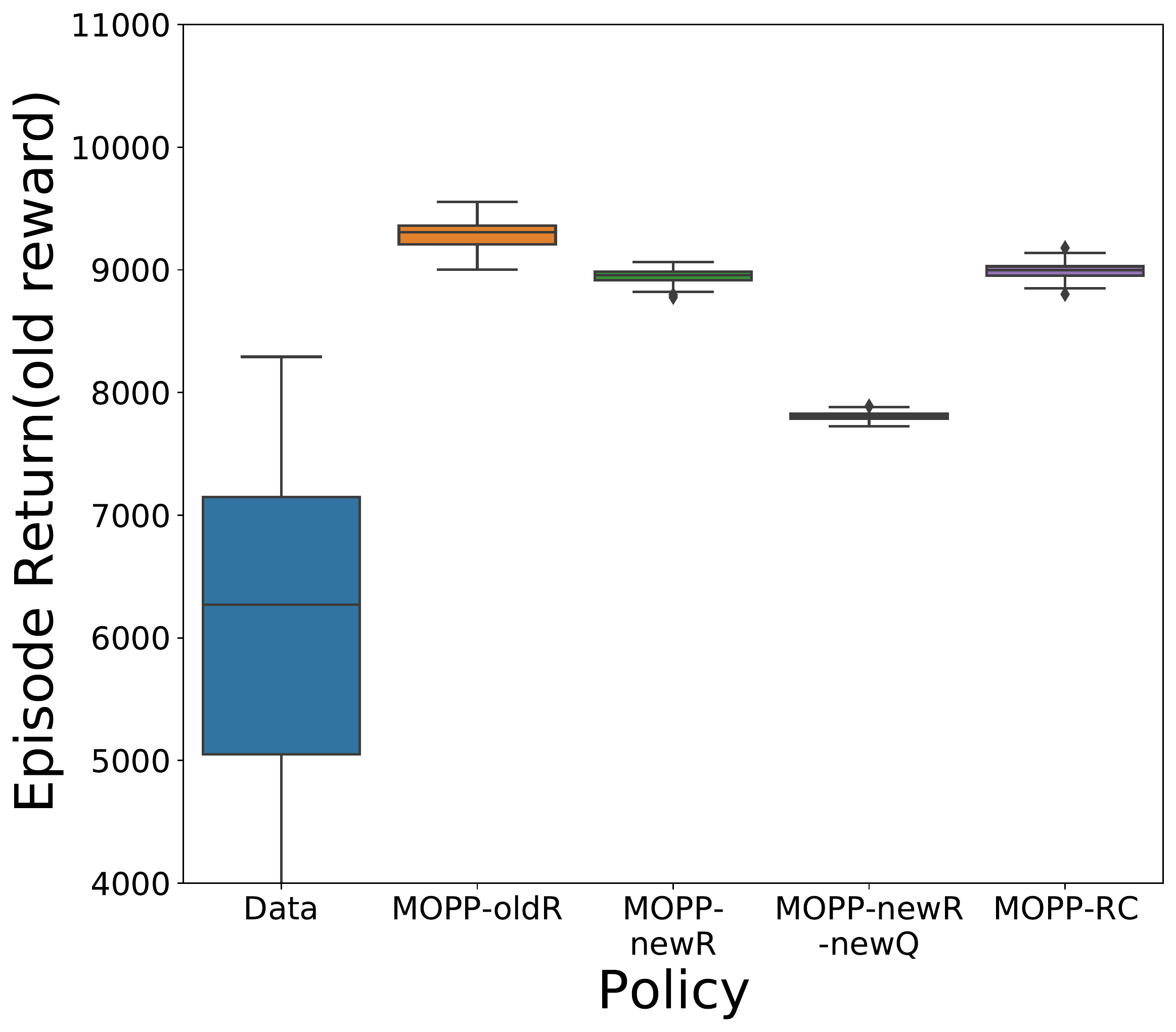}
	}\,
	\subfloat[Episode return under the new reward function]{
		\includegraphics[width=0.45\columnwidth]{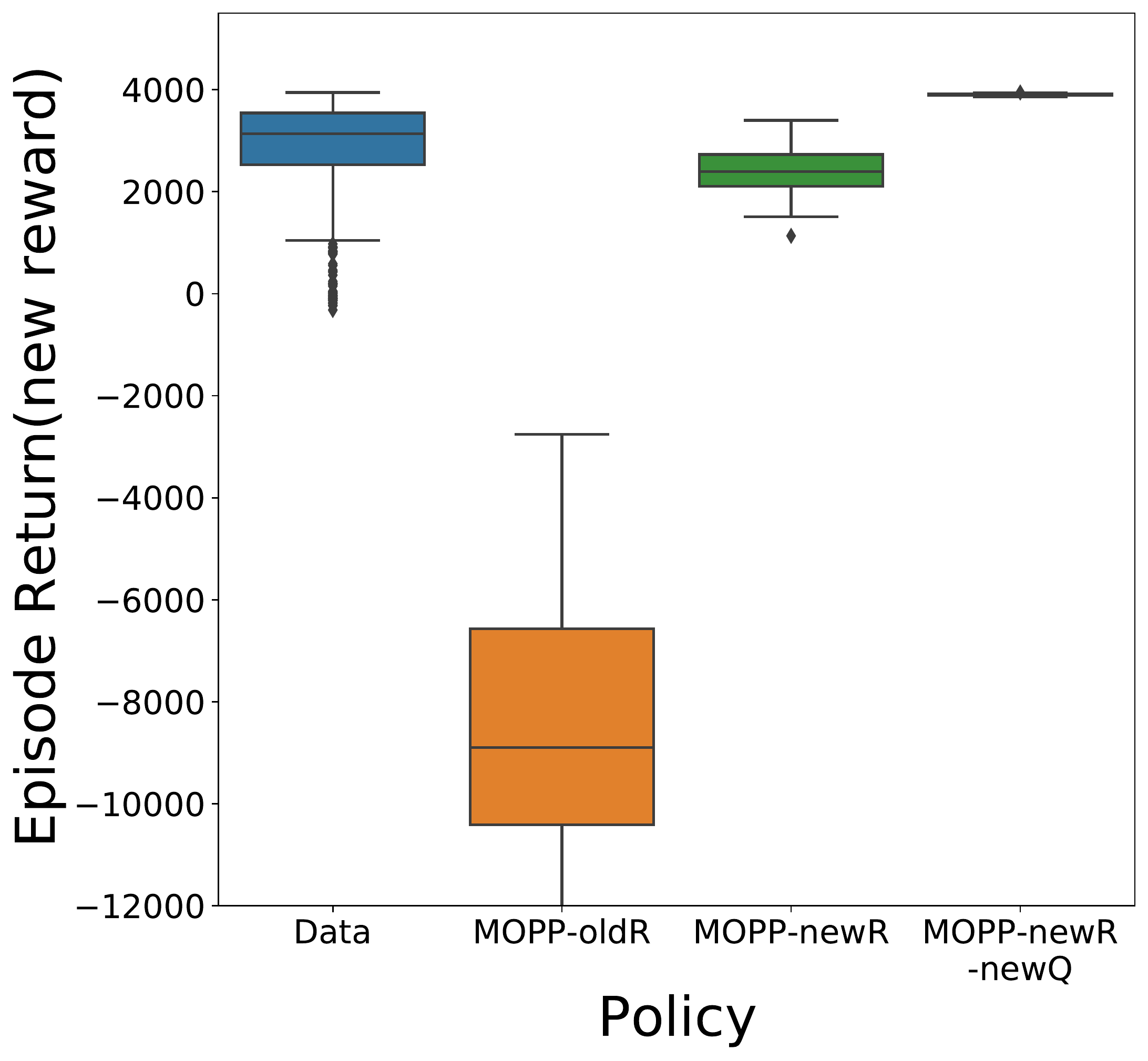}
	}
	\caption{Additional results of MOPP on \texttt{halfcheetah-} \texttt{constrained} task
	}
	\label{fig:constrained-control}
\end{figure}

Figure \ref{fig:constrained-control} presents additional results on the episode returns of MOPP and its variants under the original and new reward function of the \texttt{halfcheetah-constrained} task. Adding additional constraints to ensure safety will sacrifice the episode return measured by the original reward function. We observe that MOPP-newR and MOPP-RC can effective reduce constraint violations and have limited performance deterioration under the existence of extra constraint. Adding the constraint penalty in reward function and re-evaluating the Q-function (MOPP-newR-newQ) achieves safest policy but have substantially drop in episode return measured by the old reward function, but it still has improved episode return as measured by the new reward function.

\section{Ablation on the ADM Dynamics Model and Behavior Policy}
We also report in Table \ref{tab:adm_ablation} the additional ablation results on the impacts of using feed-forward neural networks or ADM for dynamics model and behavior policy in MOPP. For the dynamics model in MOPP, we compare the same feed-forward network structure as used in MBOP \cite{argenson2020model} (FC $f_m$) as well as our ADM dynamics model (ADM $f_m$). For the behavior policy, we report the results of the behavior policy used in MBOP (MBOP $f_b'$), the standard BC policy and the ADM behavior policy $f_b$ used in MOPP. 

As expected, the more expressive ADM dynamics model and behavior policies are found to have positive impact on the performance. Regarding the behavior policy, MBOP uses a special behavior policy that include the action of previous step as input $a_t=f_b'(s_t,a_{t-1})$, thus not directly comparable. This design will improve imitation performance under datasets generated by one or limited data generating policies, as the next action may be correlated with the previous action, but could have negative impact on high-diversity (e.g., \texttt{random} and \texttt{mixed}) or complex datasets. The more expressive ADM behavior policy $f_b$ outperforms BC policy in most tasks, except for the \texttt{random} dataset. As the expressiveness of a probabilistic model provides little help when fitting random actions. Under the same ADM dynamics model (ADM $f_m$), MOPP with ADM behavior policy $f_b$ consistently outperform the variant with BC policy. However, using ADM $f_b$ with feed-forward dynamics model (FC $f_m$) does not provide improved performance, perhaps due to less reliable trajectory rollout with the feed-forward dynamics model. Using the more expressive ADM dynamics model is observed to have larger impact on MOPP's performance as compared to the different choice of behavior policy. MOPP with ADM $f_m$ achieves improved performance in the majority of tasks under both ADM $f_b$ and BC policy. Moreover, MOPP with ADM $f_m$ and $f_b$ achieves the best overall performance.

\section{Execution Speed}
The execution speeds (control steps/second) of MOPP on the D4RL \texttt{Walker2d} and \texttt{Hopper} tasks are reported in Table \ref{tab:speed}. The tests are conducted on an Intel Xeon 2.2GHz CPU computer (no GPU involved) with simulator time included. It is observed that MOPP can easily achieve multiple controls within 1 second, which is useable for many robotics and industrial control tasks. Using longer planning horizons will increase the computation time. But we also observe in Table \ref{tab:speed} that with a moderate planning horizon (e.g., $H=8$), MOPP is already able to achieve high episode returns by incorporating the value function $V_b$ and max-Q operation with $Q_b$. The execution speed of MOPP can be further speed up by reducing the number of trajectory rollouts $N$ or use a shorter planning horizon.

In the cases when the reward function is drastically changed during system operation, to guarantee the best model performance, it is suggested to re-evaluate the Q-value function based on the new reward function. In MOPP, the Q-value function is evaluated via FQE, which is performed by simple supervised learning and computationally cheap to train. Table \ref{tab:Q-re-evaluated} presents the computation time for Q-value evaluation under different size of behavioral data for \texttt{HalfCheetah} and \texttt{Hopper} tasks. The entire computation can be finished in a relatively short time with limited resources.

\begin{table}[H]
	\centering
	\footnotesize
	\begin{tabular}{c|cl|cl}
		\toprule
		& \multicolumn{2}{|c|}{\textbf{Walker2d}}&\multicolumn{2}{|c}{\textbf{Hopper}}\\
		\midrule
		\textbf{H} & \textbf{Freq.} & \textbf{Ep. return} & \textbf{Freq.} & \textbf{Ep. return}\\ 
		\midrule
		4 & 2.69 &3885.2 $\pm$ 941.8 & 4.22 & 2539.6 $\pm$ 1051.7      \\ 
		8 & 2.13 &4032.8 $\pm$ 450.8 & 3.25 & 2847.9 $\pm$ 992.6     \\ 
		16 & 1.50 &4000.2 $\pm$ 643.8 & 2.41 & 2974.9 $\pm$ 1037.9     \\ 
		\bottomrule
	\end{tabular} 
	\caption{Execution speeds (control frequency (Hz)) and episode returns of MOPP. Models are trained on \texttt{med-expert} dataset.}
	\label{tab:speed}
\end{table}	
	
\begin{table}[H]
	\centering
	\footnotesize
	\begin{tabular}{c|c|c|c|c}
		\toprule
		& \multicolumn{2}{|c}{\textbf{HalfCheetah}}&\multicolumn{2}{|c}{\textbf{Hopper}}\\
		\midrule
		\textbf{Data size} & \textbf{1,000,000} & \textbf{200,000} & \textbf{1,000,000} & \textbf{200,000}\\ 
		\midrule
		\textbf{Time cost(min)} & 26.0 &5.2 & 24.7 & 4.9      \\ 
		\bottomrule
	\end{tabular} 
	\caption{Computation time of the Q-value function evaluation via FQE. Batch size: 512, epochs: 40. Tests are conducted on a quad-core CPU and 8 GB memory computer (no GPU involved).}
	\label{tab:Q-re-evaluated}
\end{table}

\section{Model Configurations of MOPP}
For all the D4RL MuJoCo and Adroit benchmark tasks, we use the following model configurations for MOPP.

\subsection{ADM behavior policy and dynamics model}
The ADM behavior policy $f_b$ and dynamics model $f_m$ share the same model configurations, which are set as follows:
\begin{itemize}
	\item Embedding layer: (500,)
	\item FC layers for separate dimension of output: (200, 100)
	\item Number of networks in the ensemble: 3
	\item Learning rate: 0.001
	\item Training steps: 5e+5
	\item Optimizer: Adam
\end{itemize}

\subsection{Q-value network $Q_b$}
The model configurations of $Q_b$ are set as follows:
\begin{itemize}
	\item FC layers: (500, 500)
	\item Learning rate: 0.001
	\item Training steps: 5e+5
	\item Optimizer: Adam
\end{itemize}

\end{document}